\documentclass{article}
\usepackage{PRIMEarxiv}

\usepackage[utf8]{inputenc}
\usepackage[T1]{fontenc}
\usepackage{amsmath, amssymb, amsthm, mathtools, bm}
\usepackage{graphicx}
\usepackage{svg}
\usepackage{tikz}
\usetikzlibrary{positioning,arrows.meta,arrows,shapes,fit,decorations.pathreplacing}
\usepackage{booktabs}
\usepackage{multirow}
\usepackage{multicol}
\usepackage[table,xcdraw]{xcolor}
\usepackage{mathrsfs}
\usepackage{nicefrac}
\usepackage{microtype}
\usepackage{subcaption}
\usepackage[colorinlistoftodos]{todonotes}
\usepackage{hyperref}
\usepackage{url}
\usepackage{calc} 
\usepackage{xcolor}
\usepackage{courier}
\usepackage{algorithm}
\usepackage{algpseudocode}
\algrenewcommand\algorithmiccomment[1]{\hfill\textcolor{gray}{\small\ttfamily // #1}}

\usepackage{setspace}
\usepackage{natbib}

\title{MORSE: Multi-Objective Reinforcement Learning via Strategy Evolution for Supply Chain Optimization}

\author{
  Niki Kotecha \\
  Department of Chemical Engineering \\
  Imperial College London \\
  London, SW7 2AZ \\
  \texttt{n.kotecha22@imperial.ac.uk} \\
  \And
  Ehecatl Antonio del Rio Chanona\thanks{Corresponding author} \\
  Department of Chemical Engineering \\
  Imperial College London \\
  London, SW7 2AZ \\
  \texttt{a.del-rio-chanona@imperial.ac.uk} \\
}

\begin{document}
\maketitle

\begin{abstract}
In supply chain management, decision-making often involves balancing multiple conflicting objectives, such as cost reduction, service level improvement, and environmental sustainability. Traditional multi-objective optimization methods, such as linear programming and evolutionary algorithms, struggle to adapt in real-time to the dynamic nature of supply chains. In this paper, we propose an approach that combines Reinforcement Learning (RL) and Multi-Objective Evolutionary Algorithms (MOEAs) to address these challenges for dynamic multi-objective optimization under uncertainty. Our method leverages MOEAs to search the parameter space of policy neural networks, generating a Pareto front of policies. This provides decision-makers with a diverse population of policies that can be dynamically switched based on the current system objectives, ensuring flexibility and adaptability in real-time decision-making. We also introduce Conditional Value-at-Risk (CVaR) to incorporate risk-sensitive decision-making, enhancing resilience in uncertain environments. We demonstrate the effectiveness of our approach through case studies, showcasing its ability to respond to supply chain dynamics and outperforming state-of-the-art methods in an inventory management case study. The proposed strategy not only improves decision-making efficiency but also offers a more robust framework for managing uncertainty and optimizing performance in supply chains.
\end{abstract}


\section{Introduction}

In recent years, the field of Process Systems Engineering (PSE) has experienced a significant shift towards advanced optimization techniques that are capable of handling the inherent complexities of modern systems, whether in process control \citep{linan2025trends}, sustainable supply chains \citep{qiu2024leveraging, grossmann2010scope, rangel2021machine}, or other industrial operations \citep{szatmari2024resilience}. A key challenge in PSE is the need to optimize multiple, often conflicting objectives simultaneously, such as minimizing costs, maximizing throughput, ensuring quality, and reducing environmental impact. This complexity has become particularly evident as the development of sustainable practices has emerged as a key strategic focus for many organizations, driven by the increase in regulatory requirements and pressure from consumers to adopt environmentally friendly practices. As industries strive towards making their value chains more sustainable, the integration of these principles into every aspect of their operations, particularly in decision-making, becomes essential. 

The historical roots of multi-objective optimization trace back to the work of Vilfredo Pareto in the late 19th century \citep{pareto1935mind}. Pareto efficiency, or Pareto optimality, refers to a situation where no objective can be improved without worsening another. In the context of multi-objective optimization, Pareto-optimal solutions represent a set of decisions where the trade-offs between competing objectives are balanced in the most efficient way. This foundational concept forms the basis of the modern approach to multi-objective optimization, where the goal is to generate a set of Pareto-optimal solutions for decision-makers to choose from, based on their specific priorities \citep{gunantara2018review}.

While traditional MOO method, including Linear Programming (LP) \citep{ozceylan2013fuzzy} and Evolutionary Algorithms (EAs) \citep{liao2011evolutionary}, are well-established, the increasing complexity of real-world process systems requires dynamic and adaptable solutions. This has led to the emergence of Dynamic Multi-Objective Optimization (DMOO), which seeks to optimize systems that evolve over time in response to changing environmental, operational, and economic conditions. DMOO methods adapt to evolving conditions, re-optimizing solutions as objectives shift, which is particularly important for process systems engineering where decisions must be continuously adapted to accommodate variations in process dynamics, disturbances, and uncertain operating conditions which require fast and flexible decision-making.  In such cases, the Pareto set must be recomputed as the environment changes. However, in fast-paced environments, the computational overhead involved in repeatedly solving DMOO problems can hinder fast, adaptive, real-time decision-making. 

This motivates the use of multi-objective reinforcement learning (MORL), which builds on RL principles to address multi-objective decision-making. Recent advances in MORL have demonstrated its ability to solve multi-objective decision-making problems in a series of applications in process systems engineering domains \citep{li2022multi}, inventory management \citep{qiu2024leveraging}, robotics control \citep{xu2020prediction} and energy management \citep{wu2023multi}. 

\subsection{Related Work}
Traditional Multi-Objective Optimization (MOO) approaches focus on finding a set of optimal solutions, known as the Pareto optimal, that balance trade-offs between different objectives such as cost, service level and environmental emissions. The collection of all Pareto optimal solutions forms the Pareto Front, which provides decision-makers with a range of non-dominated solutions, allowing them to choose the most appropriate solution based on their system requirements and preference. Methods such as Linear Programming (LP) \citep{ozceylan2013fuzzy}, Goal Programming \citep{dutta2015application}, and Evolutionary Algorithms (EAs) \citep{liao2011evolutionary}, including Non-dominated Sorting Genetic Algorithm II (NSGA-II) \citep{hnaien2010multi}, have been widely used to approximate the Pareto front effectively. These approaches focus on two key criteria: (1) \textit{proximity} - ensuring solutions closely approximate the true Pareto front, and (2) \textit{diversity} - ensuring solutions span a wide range of trade-offs across the objective space. MOO has been widely applied in inventory management, integrating numerous objectives such as cost efficiency, service levels, and sustainable practices \citep{aslam2010multi}. For example, \cite{determ} explored a deterministic multi-objective optimization process that targeted multi-item inventory management, emphasizing the balance between reducing shortages and controlling costs through multi-criteria decision-making techniques. Recent literature has also explored the integration of sustainability principles into inventory management paradigms, addressing energy consumption and risk of shortages across multi-storage supply chains \citep{xie2024multi, jayarathna2021multi, srivastav2016multi}. This highlights the trend towards incorporating sustainability metrics, emphasizing the need of addressing environmental issues alongside traditional inventory management strategies. Simulation-based optimization approaches have also emerged to develop frameworks that better equip decision-makers by evaluating various inventory strategies whilst accounting for multiple conflicting objectives \citep{tsai2017simulation, amodeo2009comparison, lee2005application}. However, as the number of objectives increases, multi-objective optimization becomes computationally challenging due to the curse of dimensionality, requiring more sophisticated methods such as decomposition-based algorithms \citep{abbassi2022elitist} and indicator-based approaches \citep{yang2016multi}. 

In dynamic environments, where objectives change over time, MOO techniques can be extended to dynamic multi-objective optimization (DMOO) methods. DMOO methods adapt to evolving conditions, re-optimizing solutions as objectives shift, which is particularly important in the context of supply chains, where external disruptions and market fluctuations require fast and flexible decision-making.  In such cases, the Pareto set must be recomputed as the environment changes. However, in fast-paced supply chains, the computational overhead involved in repeatedly solving DMOO problems can hinder fast, adaptive, real-time decision-making \citep{helbig2016key}. 

\subsubsection{Multi-Objective Reinforcement Learning (MORL)}
To address these shortcomings, multi-objective reinforcement learning (MORL) has emerged as a promising alternative that integrates the principles of DMOO with RL \citep{hayes2022practical}. Recent advances in MORL have demonstrated its ability to solve multi-objective decision-making problems in a series of applications such as robotics control \citep{tran2023two}, cloud computing \citep{khan2024dynamic} and energy management \citep{wu2023multi}. 

\paragraph{Scalarization-Based Approaches} 
In Multi-Objective Reinforcement Learning (MORL), the challenge lies in optimizing multiple conflicting objectives simultaneously. Early approaches often utilized scalarization techniques, where multiple objectives are combined into a single utility function using weighted sums \citep{van2013scalarized, hayes2022practical}. The utility function represents the combined measure of performance across all objectives, which simplifies the multi-objective problem into a single-objective framework, making it compatible with standard reinforcement learning (RL) algorithms. However, scalarization methods suffer from several limitations. They assume a predefined weight for each objective, which can lead to poor performance when the Pareto front is highly non-convex or when preferences change dynamically over time. Additionally, scalarization struggles to maintain a diverse set of policies, especially when objectives conflict, since it focuses on a single trade-off between objectives \citep{abels2019dynamic}.

\paragraph{Pareto-Based Multi-Policy Methods}
To address these challenges, Pareto-based methods, also known as multi-policy methods, were introduced \citep{hayes2022practical}. These methods aim to approximate the Pareto front by learning and maintaining a diverse set of policies, each corresponding to a different trade-off between objectives. This allows for better coverage of the solution space, particularly in complex, dynamic environments.

Multi-policy methods in MORL focus on simultaneously learning and storing multiple Pareto-optimal policies. Unlike single-policy approaches that aim to find one solution, multi-policy methods aim to find a set of diverse solutions that balance conflicting objectives. These methods are typically divided into inner-loop and outer-loop approaches \citep{hayes2022practical}.

Inner-loop methods modify the learning algorithm to identify and store multiple policies during a single training process. For instance, Pareto-Q-Learning (PQL) \citep{van2014multi} and PQ-learning \citep{ruiz2017temporal} extend traditional Q-learning by maintaining multiple Pareto-optimal Q-values for each state-action pair. These methods also incorporate pruning techniques to eliminate dominated policies, ensuring that only the most valuable policies are retained. This approach allows for efficient learning and storage of policies for different trade-offs without the need for multiple training cycles.

On the other hand, outer-loop methods involve running a separate training process for each utility function or combination of objectives \citep{parisi2014policy}. This process iterates through different parameter settings of the utility function, which could be different weightings of the objectives, and re-runs the learning algorithm for each setting. While this approach can be effective, it is often inefficient because it requires multiple runs of the algorithm, potentially re-learning similar policies or missing diverse solutions due to the discrete nature of the search \citep{hayes2022practical, roijers2015point}.

Further advancements include Multi-Objective Fitted Q-Iteration (MOFQI) \citep{castelletti2012tree}, which adapts the Fitted Q-Iteration algorithm \citep{ernst2005tree} to the multi-objective domain by incorporating scalarization weights directly into the state representation. This allows for the approximation of the optimal Q-function across all possible scalarization weights, enabling the algorithm to handle multiple objectives within a single training process. Additionally, multi-objective extensions of Monte Carlo Tree Search (MCTS) have been applied, based on hypervolume metrics \citep{wang2012multi, wang2013hypervolume} or Pareto-dominance \citep{weng2020pareto} to guide the search towards optimal trade-offs and improve the efficiency of the algorithm in exploring the solution space.

Compared to outer-loop methods, inner-loop multi-policy approaches are generally more efficient because they allow multiple policies to be learned in parallel within a single training process, without the need to re-train for each utility function. 

\paragraph{Population-Based Evolutionary Reinforcement Learning}
Moreover, the widely studied population-based evolutionary methods are well-suited for multi-policy methods due to their ability to explore a diverse set of solutions \citep{branke2008multiobjective, deb2002fast, murata1995moga}. Evolutionary algorithms have been combined in a hybrid approach, where global search methods are used for broad exploration, and local refinement strategies, such as hill-climbing \citep{soh2011evolving} or policy-gradient methods \citep{xu2020prediction}, are employed to fine-tune policies and optimize them for specific areas of the solution space.

Therefore, multi-policy methods in MORL learn a Pareto set of policies where each policy corresponds to a behavioral strategy that offers a different trade-off between objectives, similar to the Pareto set of solutions in traditional MOO. However, unlike static MOO, these policies are learned through exploration in dynamic and uncertain environments. By integrating the dynamic and adaptive capabilities of reinforcement learning, MORL provides a data-driven framework that can better manage evolving objectives and real-time, fast decision-making in supply chains.

\subsection{Motivation}
In this work, we propose leveraging the capabilities of multi-objective evolutionary algorithms to search the parameter space of the neural network policies. This leads to a dense, high-quality set of control policies that trade-off multiple conflicting objectives. In summary, the key contributions of this work are as follows: 
\begin{itemize}
    \item We develop a novel framework that integrates multi-objective evolutionary algorithms with reinforcement learning, enabling the simultaneous optimization of competing objectives within a complex environment.
    \item We introduce a Conditional Value-at-Risk (CVaR) approach for distributional multi-objective RL, which provides a framework to account for the worst-case distributional risks.
    \item We benchmark our approach against state-of-the-art multi-objective reinforcement learning methods, demonstrating superior performance and adaptability in dynamic and uncertain environments.
\end{itemize}

The rest of this paper is organized as follows: Section \ref{background} provides the background on reinforcement learning and evolutionary strategies for reinforcement learning, Section \ref{method} describes the proposed multi-objective evolutionary algorithm based reinforcement learning framework in more detail. Section \ref{result} discusses the simulation and experimental results on a 3-dimensional case study. Section \ref{robust} investigates the robustness of the learned policies using a Conditional Value-at-Risk (CVaR) formulation. Section \ref{benchmark} compares the performance of the proposed method against established benchmarks. Finally, Section \ref{conc} summarizes the paper and provides an outlook for future work. 

\section{Preliminaries} \label{background}
We first provide a background into the components that contribute towards our methodology including single agent reinforcement learning and evolutionary strategies. Readers already familiar with these topics may skip directly to Section \ref{method}. 

\subsection{Reinforcement Learning}
\begin{figure}
    \centering
    \includegraphics[width=0.5\linewidth]{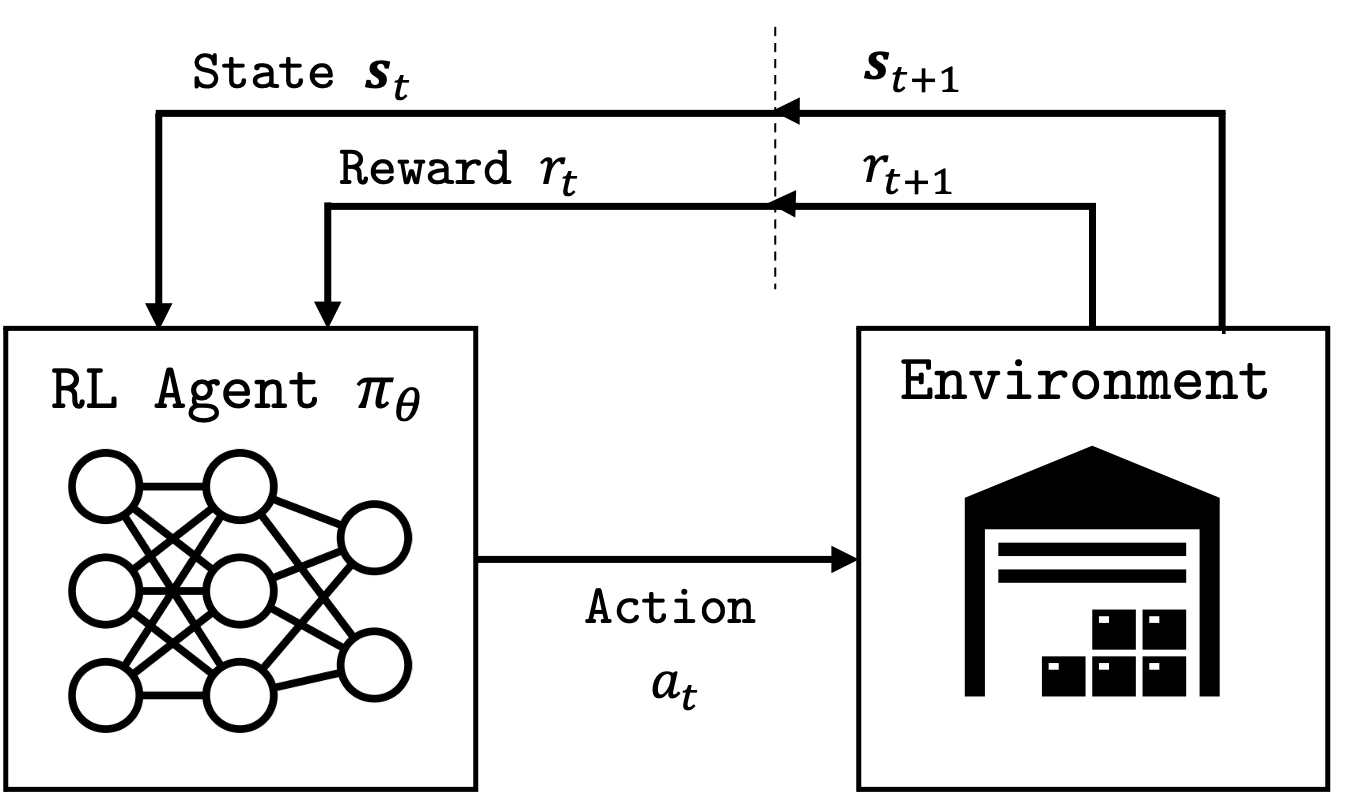}
    \caption{Illustration of the reinforcement learning process. The agent observes the environment, takes an action based on its policy, receives a reward, and transitions to a new state, iterating through this cycle to improve its policy over time.}
    \label{fig:rl}
\end{figure}

In single agent reinforcement learning, the agent aims to learn an optimal policy by interacting with the environment and learning through trial-and-error as shown in Figure \ref{fig:rl}. In RL, the agent observes the current state $\mathbf{s}_t \in \mathcal{S}$, chooses an action $\mathbf{a}_t \in \mathcal{A}$ and transitions into the next state $\mathbf{s}_{t+1} \in \mathcal{S}$. For a deterministic policy $\pi$, the agent takes action $\mathbf{a}_t = \pi(\mathbf{s}_t)$, and the transition to the next state is governed by the state transition probability function $P(\mathbf{s}_t, \mathbf{a}_t, \mathbf{s}_{t+1})$ which reflects the probability of transitioning to $\mathbf{s}_{t+1}$ given $\mathbf{s}_t$ and $\mathbf{a}_t$. For a stochastic policy $\pi$, the action $\mathbf{a}_t$ is sampled from a conditional probability distribution  $\mathbf{a}_t \sim \pi(\cdot|\mathbf{s}_t)$. The overall transition probability is from $\mathbf{s}_t$ to $\mathbf{s}_{t+1}$ under policy $\pi$ is computed by averaging over all possible actions, weighted by the probability of each action under the policy. After taking the action $\mathbf{a}_t$, the agent receives a reward $r_{t+1} = \mathcal{R}(\mathbf{s}_t, \mathbf{a}_t, \mathbf{s}_{t+1})$. 

The agent finds an optimal policy $\pi^*$ by maximizing the expected sum of rewards over a time horizon defined as: 
\begin{align}
    J(\pi) &= \mathbb{E}_{\pi}\left[ \sum_{t=0}^{T} \gamma^t r_t(\mathbf{s}_t, \mathbf{a}_t)\right] \\
    \pi^* &= \arg\max_{\pi}J(\pi)
\end{align}

In Deep RL, the policy is parameterized by a neural network, where policy $\pi \approx \pi_\theta$ and $\mathbf{\theta} \in \Theta \subseteq \mathbb{R}^{n_\theta}$ represents the weights of the neural network and $\Theta$ is the parameter space. This neural network approximates the policy, allowing the agent to handle high-dimensional state and action spaces that would be intractable for traditional RL methods.

There are two major classes of model-free algorithms in RL: policy-based and value-based methods. Value-based methods focus on learning a value function, which estimates the expected return for a given state (or state-action pair). The value function $V^{\pi}(\mathbf{s}_t)$ measure how \textit{good} it is for the agent to be in a given state $\mathbf{s}_t$ under the policy $\pi$, quantified by the expected return:
\begin{equation}
    V^{\pi}(\mathbf{s}_t) = \mathbb{E}_{\pi}\left[J(\pi)|\mathbf{s}_0=\mathbf{s}\right]
\end{equation}

The action-value function (also called the Q-function), measures the expected return from a given state $\mathbf{s}_t$ and taking action $\mathbf{a}_t$ under the policy $\pi$, given by:
\begin{equation}
    Q^{\pi}(\mathbf{s}_t, \mathbf{a}_t) = \mathbb{E}_{\pi}\left[J(\pi)|\mathbf{s}_0=\mathbf{s}, \mathbf{a}_0 = \mathbf{a}\right]
\end{equation}

Therefore, value-based methods aim to maximize the value or action-value function by learning an optimal policy that selects actions based on the highest value, such as in Deep Q-Networks (DQN). 

On the other hand, policy-based methods directly optimize the policy itself without explicitly calculating the value function. These methods parameterize the policy as a function $\pi_\theta$ and the goal is to optimize the parameters $\mathbf{\theta}$ to maximize the expected cumulative reward $J(\pi_\theta)$. A widely used class of policy-based methods are policy gradient methods which use gradient ascent to update the policy parameters based on sampled trajectories. Examples of policy gradient methods include REINFORCE. The policy parameters are updated by using stochastic gradient ascent on the objective function with a scalar learning rate $\alpha$:

\begin{equation}
    \mathbf{\theta} \leftarrow \mathbf{\theta} + \alpha \nabla_{\mathbf{\theta}} \hat{J}(\mathbf{\theta})
\end{equation}

Policy gradient methods offer several advantages as they are suitable for high dimensions and are particularly effective in environment with continuous action spaces. However, policy gradient methods often rely on sampling rollouts (trajectories) to estimate gradients. Due to the stochastic nature of the environment, these samples can vary greatly and hence lead to a high variance in gradient estimates, causing the optimization to converge to locally optimal policies.

\subsection{Evolutionary Strategies for Reinforcement Learning}
One alternative to solving RL problems is using black-box optimization techniques like Evolutionary Strategies. Evolution Strategies (ES) are population-based search methods inspired by the process of natural evolution.  Instead of relying on gradient-based updates, evolutionary strategies explore the parameter space of the policy by maintaining a population of candidate solutions (policies). In other other words, we are optimizing a function $\pi(\theta)$ with respect to the input vectors $\theta$ (weights of the policy) but we make no assumptions about the structure of the function $\pi$ except that we can evaluate the function by interacting with the environment.

In the context of reinforcement learning, evolutionary strategies are used to optimize the parameters $\mathbf{\theta}$ of a policy $\pi(\mathbf{\theta})$ without relying on gradients, by evaluating a population of policies represented by a set of parameter vectors $\Theta = \{\mathbf{\theta}_1, \mathbf{\theta}_2, \cdots, \mathbf{\theta}_{n_{\pi}}\}$, where each $\mathbf{\theta}_i \in \mathbb{R}^{n_\theta}$ defines a distinct policy in the population, $n_\theta$ is the number of parameters in the policy network (e.g., the total number of weights and biases) and $n_\pi$ is the total number of policies in the population. Each policy $\pi(\mathbf{\theta})$ parameterized by $\mathbf{\theta}$ is executed over multiple episodes to obtain its expected cumulative return $J(\mathbf{\theta})$. The estimated performance of each policy is then used to update the parameters $\mathbf{\theta}$ towards those that maximize $J(\mathbf{\theta})$. 

Evolutionary strategies for reinforcement learning has shown promising results in optimizing policy parameters \citep{sigaud2023combining, khadka2018evolution}. Instead of relying on gradient information, ES-RL evaluate policy performance by sampling multiple trajectories and directly updating the parameters towards those that lead to higher performance. There's also fewer hyperparameters to tune compared to gradient-based methods and they are less likely to get stuck in local optima as they are population-based methods so search "globally" rather than relying on stochastic estimates of gradients to optimize the parameters. The advantages of ES-RL methods have led to significant research in this space. One notable work was OpenAI's algorithm that leveraged on the strengths of evolutionary strategies which showed comparable performance compared to other policy gradient methods on well known environments such as MuJoCo and Atari \citep{salimans2017evolution}.

\section{Methodology} \label{method}
This section presents the proposed multi-objective reinforcement learning framework which integrates traditional reinforcement learning with multi-objective evolutionary strategies. The methodology section also covers the problem definition and agent design. 

\subsection{Problem Definition} \label{sec:case_study}
Our multi-echelon, multi-product inventory management model as shown in Figure \ref{fig:im} integrates three cumulative objective functions throughout the time horizon: Maximize the profit across all nodes, minimize the transportation emission across all nodes, minimize the lead time across all nodes. The resulting multi-objective optimization problem can be formulated as follows:
 \begin{figure}[h]
     \centering
     \includegraphics[width=0.6\linewidth]{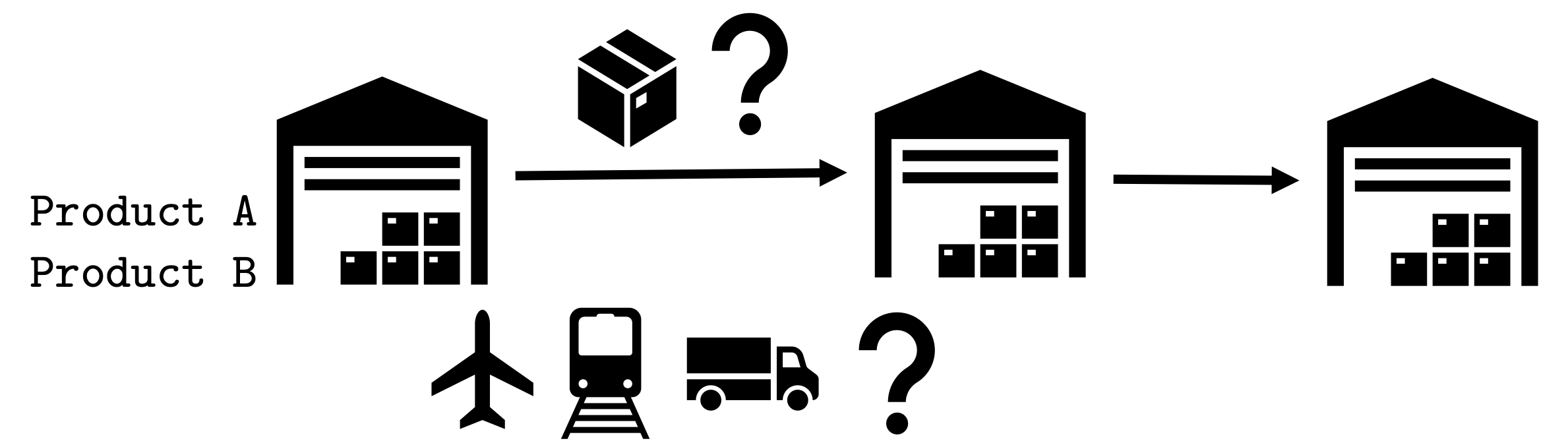}
     \caption{Schematic representation of the inventory management system setup, illustrating key components and their interactions.}
     \label{fig:im}
 \end{figure}
 
\begin{align*}
\max & \sum_{m=1}^{m_{\text{tot}}
}\sum_{p=1}^{p_{\text{tot}}
} \sum_{t=1}^{t_{\text{tot}}}
 \mathbf{P}_s^{m,p} s_r^{m,p}[t] - \mathbf{C}^{m,p} o_r^{m,p}[t] - \mathbf{T}_{r}^{m,p}(z_r^{m,p}[t]) L^{m_u} o_r^{m,p}[t] - \mathbf{I}^{m,p} i^{m,p}[t] - \mathbf{B}^{m,p} b^{m,p}[t]  & &  \\
\min & \sum_{m=1}^{m_{\text{tot}}
}\sum_{p=1}^{p_{\text{tot}}
} \sum_{t=1}^{t_{\text{tot}}} e^{m}(z_r^{m,p}[t]) L^{m_u} o_r^{m,p}[t] & &  \\
\min & \sum_{m=1}^{m_{\text{tot}}
}\sum_{p=1}^{p_{\text{tot}}
} \sum_{t=1}^{t_{\text{tot}}} \tau_r^{m,p}(z_r^{m,p}[t]) && \\
&\text{subject to:} \nonumber \\
& i^{m,p}[t] = i_0^{m,p}[t] - s_r^{m,p}[t] + a_r^{m,p}[t], \quad \forall m, \forall p, \forall t, \notag \\
& b^{m_d,p}[t] = b_0^{m_d,p}[t] - s_r^{m_d,p}[t] + d_r^{m_d,p}[t], \quad \forall m, \forall p, \forall d \in D_m, \notag \\
& s_r^{m_d,p}[t] \leq b_0^{m_d,p}[t] + d_r^{m_d,p}[t], \quad \forall m, \forall p, \forall t, \forall d \in D_m, \notag \\
& s_r^{m,p}[t] \leq i_0^{m,p}[t] + a_r^{m,p}[t], \quad \forall m, \forall p, \forall t, \notag \\
& a_r^{m,p}[t] = s_r^{m_u,p}[t-\tau_r^m], \quad \forall m \neq 1, \forall p, \forall t \geq \tau_r^m, \notag \\
& a_r^{1,p}[t] = s_r^{1,p}[t-\tau_r^1], \quad \forall p, \forall t \geq \tau_r^1, \notag \\
& d_r^{m_d,p}[t] = o_r^{d,p}[t], \quad \forall m, \forall p, \forall d \in D_m, \notag \\
& d_r^{m,p}[t] = c^{m,p}[t], \quad \forall m \in C, \forall p, \forall t, \notag \\
& o_r^{m,p}[t] \leq \mathbf{o}_{r_{\text{max}}}^m, \quad i^{m,p}[t] \leq \mathbf{i}_{\text{max}}^m, \quad \forall m, \forall p, \forall t. \notag
\end{align*}
\normalsize
The goal is to ascertain the optimal action for each node \( m\) and each product \(p\) during each time period \( t \) spanning over a total of \( t_{\text{tot}}
 \) time periods within a discrete-time setup. 

$s_r^{m,p}[t]$ is the amount of goods shipped to a downstream node (or customers); $o_r^{m,p}[t]$ is the re-order quantity; $d_r^{m,p}[t]$is the demand from downstream node(s); $a_r^{m,p}[t]$ is the acquisition at the current time step; $c^{m,p}[t]$ corresponds to customer demand; $i^{m,p}[t]$ and $b^{m,p}[t]$ are the on-hand inventory level and backlog at the end of a time period; $i_0^{m,p}[t]$ and $b_0^{m,p}[t]$ denote the initial on-hand inventory level and backlog; $\tau_r^{m}$ is lead time; $L^{m_u}$ represents the distance from node $m$ to its upstream supplier and $z_r^{m,p}[t] \in \{1,2,3,\cdots\}$ represents a discrete integer categorical variable representing the transportation type (e.g., $1=\text{truck},2=\text{rail},3=\text{air}, \cdots$). 

$\mathbf{P}_s, \mathbf{C}, \mathbf{T}_r, \mathbf{I}, \mathbf{B} \in \mathbb{R}^{m_{\text{tot}} \times p_{\text{tot}}}$ are matrices of cost coefficients - selling price, cost of re-order, transportation, stock, backlog, respectively; $\mathbf{e} = (e_1, e_2, \dots, e_M)^\top \in \mathbb{R}^{m_{\text{tot}}}$ is unit transportation emission; $\mathbf{o}_{r_{\max}} = (o_{r_{\max}}^1, \ldots, o_{r_{\max}}^{m_{\text{tot}}})^\top \in \mathbb{R}^{m_{\text{tot}}}$ and $\mathbf{i}_{\max} = (i_{\max}^1, \ldots, i_{\max}^M)^\top \in \mathbb{R}^{m_{\text{tot}}}$ represent the maximal re-order amount and node storage capacity, respectively. ${m_{\text{tot}}}$ represents the total number of nodes and ${p_{\text{tot}}}$ represents the total number of products. The subscript $u$ refers to the upstream node, $d$ denotes the downstream node.

In our supply chain environment, we incorporate exogenous uncertainty through two primary stochastic elements: lead time and customer demand. Lead time is modeled as a Poisson random variable with parameter $\lambda_\tau$. Customer demand is modeled in two separate experimental setups to examine different stochastic scenarios. In the first experiment, demand arrivals follow a standard Poisson process with rate $\lambda_0$, capturing random variability without seasonal effects. In the second experiment, demand is modeled as a non-stationary Poisson process with a sinusoidal mean function to reflect periodic fluctuations:
\[
\lambda_d[t] = \lambda_0 \left( 1 + A_s \sin(2\pi f t + \phi) \right)
\]
where $\lambda_0$ is the baseline demand rate, $A_s$ is the amplitude of seasonal fluctuations, $f$ is the frequency of the seasonal cycle, $\phi$ represents the phase shift and $t$ represents time. 

Therefore, the probability of observing exactly $k$ events in a fixed time period is given by: 
\begin{equation}
    P(X=k)=\frac{(\lambda)^ke^{-\lambda}}{k!}
\end{equation}
where $X$ represents a random variable (e.g., number of of events such as demand arrivals or lead time occurrences), $k \in \mathbb{N}_0$ is the number of events and $\lambda$ is the rate parameter for the specific Poisson process. 

\subsection{Inventory Control Agent}
\begin{figure}
    \centering
    \includegraphics[width=\linewidth]{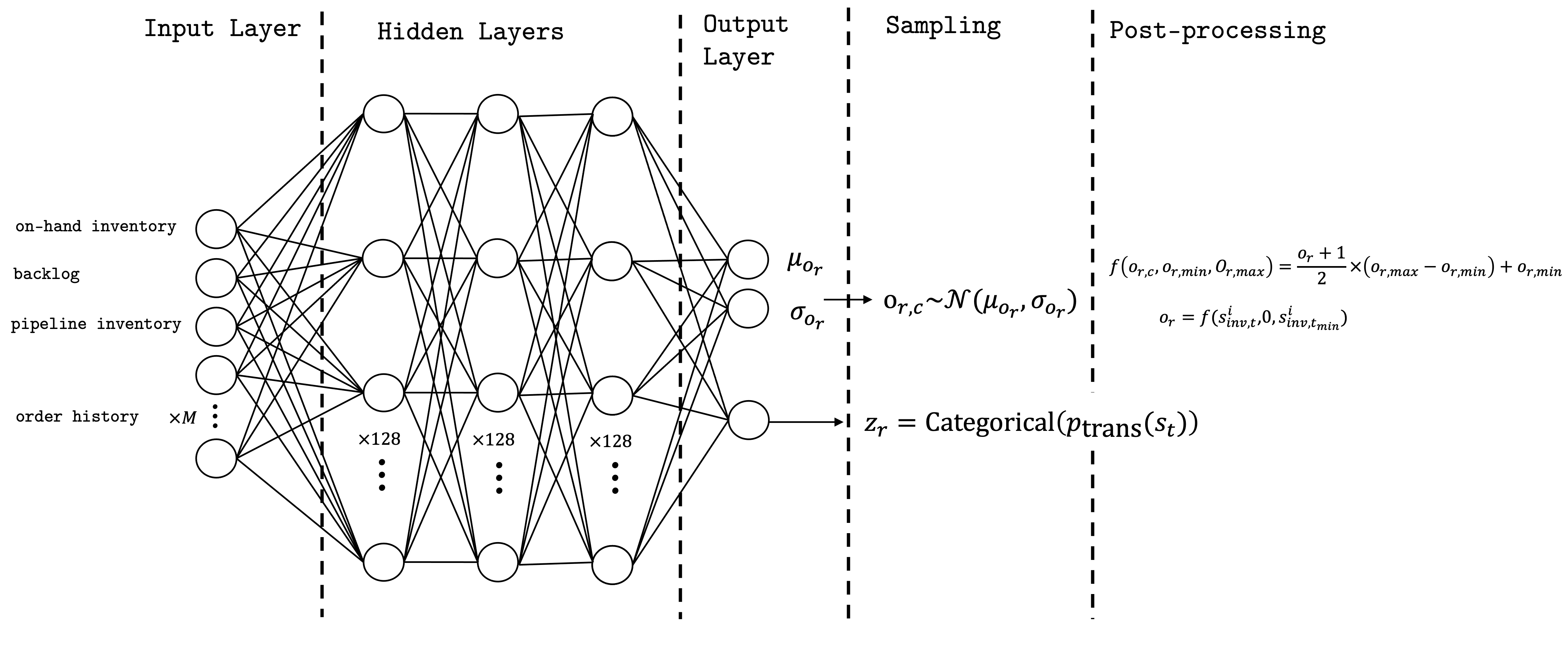}
    \caption{Architecture of the inventory control agent. The neural network takes the state space as input and outputs two actions: (1) a continuous order replenishment decision, sampled from a Gaussian distribution, followed by a post-processing step to map it to a feasible quantity, and (2) a discrete transportation mode selection, based on the highest probability from a softmax distribution over available modes.}
    \label{fig:ic_agent}
\end{figure}

The Inventory Control (IC) Agent consists of a deep neural network policy which takes the observed states $s_t$ as inputs and outputs the actions $\mathbf{a}_t$ at each time step as shown in Figure \ref{fig:ic_agent}. In the state space, a new variable is introduced, $\mathbf{V}[t]$, which is the pipeline inventory equal to the sum of order replenishment that has not yet arrived at the node from other upstream nodes. Therefore, the state is defined with the on-hand inventory,  $\mathbf{I}[t]$, pipeline inventory, $\mathbf{V}[t]$ backlog, $\mathbf{B}[t]$, demand history and order history up to $n_t$ time steps in the past where $n_t$ is a hyperparameter.  
\begin{equation}
    \mathbf{s}_t = \left[\mathbf{I}[t], \mathbf{V}[t], \mathbf{B}[t], \mathbf{O}_r[t-1],\cdots, \mathbf{O}_r[t-n_t], \mathbf{D}_r[t-1],\cdots, \mathbf{D}_r[t-n_t]\right]
\end{equation}
Despite breaking the Markovian property, incorporating historical data is a well-established approach to augment the observation space in partially observable environments, which are common in supply chain management systems due to their inherent stochastic nature. While Recurrent Neural Networks (RNNs) are capable of handling sequential data and capturing temporal dependencies, we chose not to use them here for the sake of simplicity and practicality in training. Instead, we opted for a fixed window of past observations, which allows for a more straightforward implementation while still preserving the relevant historical context. It is important to note that including historical data introduces a violation of the Markov property, which states that the future state of the system should only depend on the current state, not on the sequence of events that preceded it. However, in real-world decision-making processes, perfect Markovian behavior is rare. Thus, augmenting the observation space with historical data is a practical and well-recognized method in partially observable environments \citep{liu2022partially, marlmousa, uehara2022provably, marlgnn}.

For each node and each product at each time step in the simulation, the policy outputs two key actions: \\
\textbf{Order Replenishment}: While order replenishment is typically modeled as a discrete decision in most inventory management systems, we model it as a continuous action within the range $[-1,1]$ for scalability to a wide range of possible order sizes \citep{marlgnn}. The policy, parameterized by a neural network, outputs the mean and standard deviation of a Gaussian distribution, from which we sample the continuous action \footnote{While our method uses a stochastic policy, this is not strictly necessary in the context of evolutionary algorithms. Exploration is already handled at the population level through random initialization, mutation, and crossover. The inclusion of a stochastic policy aligns with common practice in reinforcement learning, but a deterministic policy would likely yield similar results. This reflects a modeling choice rather than a requirement, and highlights a subtle design nuance.}. This approach allows us to scale the order sizes continuously while avoiding the combinatorial explosion problem often seen in discrete and mixed-integer optimization problems, while still approximating the behavior of a discrete system. The policy for order replenishment, denoted as $\pi_{\theta_{o_r}}$, is a Gaussian distribution from which we sample the action for replenishment: 
    \begin{equation} 
    o_{r,c}\sim \mathcal{N}(\mu_{o_r}, \sigma_{o_r}^2) 
    \end{equation} 
    where $\mu_{o_r}$ is the mean and $\sigma_{o_r}$ is the standard deviation output by the network and subscript $c$ denotes the unscaled value. After sampling the action, a min-max scaling step is applied within the environment to map this continuous value to a feasible replenishment quantity.
    \begin{align}
        f(\mathbf{o}_r, \mathbf{o}_{r,min}, \mathbf{o}_{r,max}) &= \frac{\mathbf{o}_r+1}{2}\times(\mathbf{o}_{r,max}-\mathbf{o}_{r,min}) + \mathbf{o}_{r,min} \\
        \mathbf{o}_r &= f(\mathbf{o}_{r,c}, 0, \mathbf{o}_{r_{max}})
    \end{align} \\
\textbf{Transportation Mode}: A discrete action that represents the mode of transportation selected for product movement, such as air, rail, or truck. This part of the policy outputs a categorical distribution over the available transportation modes. The policy for transportation mode, denoted as $\pi_{\theta_{z_r}}$, is a categorical distribution over the modes of transportation:
\begin{equation}
    z_r \sim \pi_{\mathbf{\theta}_{z_r}}(\cdot \mid \mathbf{s}_t) \quad \text{where }\pi_{\mathbf{\theta}_z}(z \mid \mathbf{s}_t)= \text{Categorical}(p_{\text{trans}}(\mathbf{s}_t))
\end{equation}
where $\pi_{\theta_z}(z \mid \mathbf{s}_t)$ is the policy distribution over the modes of transportation,  \(p_{\text{trans}}(\mathbf{s}_t)\) is the probability vector output by the policy, representing the likelihood of each transportation mode given state $s_t$ over the transportation modes and $z_r \in \{0,1,2,\cdots,n_z\}$ is the selected transportation mode, where $n_z$ is the number of possible transportation modes.

Mathematically, the overall policy $\pi_\theta$, parametrized by $\theta$, combines these two components, producing both the continuous action for order replenishment and the discrete action for transportation mode. We express this as:
\begin{equation}
    \pi_{\theta} = (\pi_{\theta_{o_r}},\pi_{\theta_{z_r}})
\end{equation}

This sequential decision-making problem is formulated as a Multi-Objective Markov Decision Process, which can be defined as a tuple $\left\langle \mathcal{S},\mathcal{A},\mathcal{T},\gamma, \rho_0, \mathbf{R} \right\rangle$ where $\mathcal{S}$ is the state space, $\mathcal{A}$ is the action space, $\mathcal{T}$ is the probability transition function where $\mathcal{T}: \mathcal{S} \times \mathcal{A} \times \mathcal{S} \rightarrow [0,1]$, $\gamma \in [0,1]$ is the discount factor, $\rho_0 : \mathcal{S} \rightarrow [0,1]$ is a probability distribution over initial states and $\mathbf{R}: \mathcal{S} \times \mathcal{A} \times \mathcal{S} \rightarrow \mathbb{R}^{n_f}$ is a vector valued reward function where $n_f \geq 2$ is the number of objectives. The vector-valued reward function $\mathbf{R}$ is one of the differences between single-objective RL an multi-objective RL. Finally, a policy $\pi: \mathcal{S} \rightarrow \mathcal{A} \in \Pi$, maps states to actions where $\Pi$ is a set of all the possible policies. Another notable difference between multi-objective and single-objective MDPs is the vector valued value function, $\mathbf{V}^\pi \in \mathbb{R}^{n_f}$ which is conditioned on the number of objectives and is defined as $\mathbf{V}^\pi (\mathbf{s}) = \mathbb{E}_\pi \left[ \sum_{t=0}^{\infty} \gamma^t \mathbf{r}_{t+1}\right]$ where $\mathbf{r}_{t+1} = \mathbf{R}(\mathbf{s}_t, \mathbf{a}_t, \mathbf{s}_{t+1})$. A note that due to the multi-objective nature of the definition, it is possible to encounter a situation where for objectives $j$ and $k$ where $j,k \in \left\{1,2, \dots, n_f\right\}$ and policies $\pi$ and $\pi'$ where $\pi, \pi' \in \Pi$, both of the following inequalities can hold true: 
\begin{align}
    V_{j}^\pi (\mathbf{s}) &> V_{j}^{\pi'}(\mathbf{s}) \quad \text{(Policy } \pi \text{ is better for objective } j) \\
    V_{k}^\pi (\mathbf{s}) &< V_{k}^{\pi'}(\mathbf{s}) \quad \text{(Policy } \pi' \text{ is better for objective } k)
\end{align}

This indicates that while policy $\pi$ outperforms policy $\pi'$ in objective $j$, it underperforms in terms of objective $k$. This illustrates the concept of trade-offs in multi-objective reinforcement learning, where optimizing for one objective can lead to suboptimal performance in another. Therefore, the agent must make decisions based on the relative importance of each objective. 

\subsection{\textbf{MORSE} - \textbf{M}ulti-\textbf{O}bjective \textbf{R}einforcement learning via \textbf{S}trategy \textbf{E}volution}

\begin{figure}
    \centering
    \includegraphics[width = 0.8\linewidth]{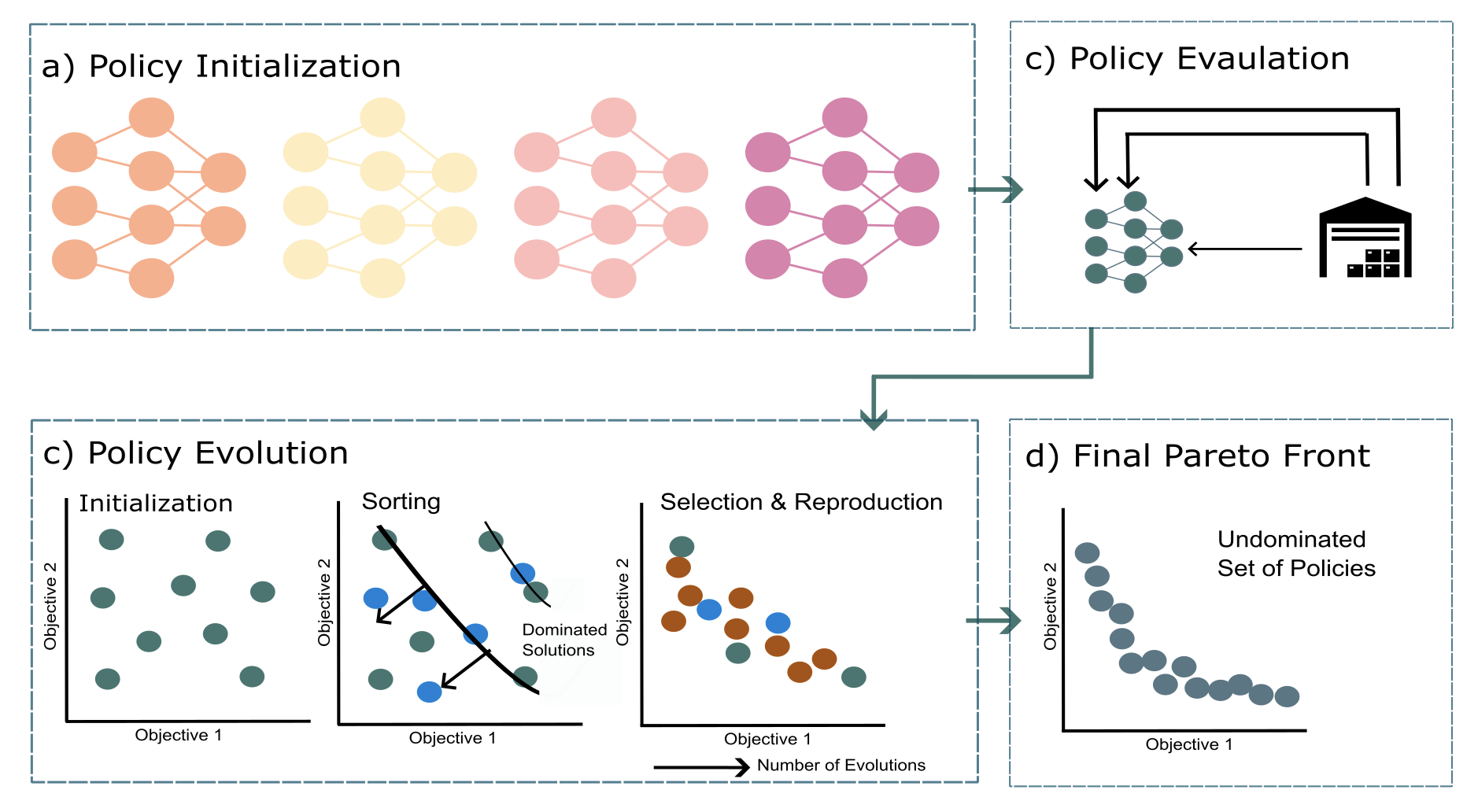}
    \caption{Overview of the MORSE: Multi-Objective Reinforcement Learning via Strategy Evolution showing: (a) Initialization of a diverse population of policies with randomly sampled parameters. (b) Evaluation of each policy across multiple objectives using sampled episodic returns. (c) Evolutionary process guided by non-dominated sorting and crowding distance to promote convergence and diversity. (d) Final Pareto front representing the set of non-dominated policies that balance trade-offs among competing objectives.}
    \label{fig:morse}
\end{figure}

Due to the aforementioned benefits of evolutionary strategies, the proposed methodology leverages multi-objective evolutionary strategies (MOEA) for multi-objective reinforcement learning. In this work, we leverage MOEA to directly optimize the parameter space of the policies, building a Pareto set of policies rather than a Pareto set of solutions, as is common in traditional multi-objective optimization methods. This approach provides the decision maker with a diverse set of adaptable and dynamic policies, facilitating rapid decision-making in complex environments. The proposed methodology is shown in Algorithm 1 and Figure \ref{fig:morse} and the steps can be described as follows: 

\begin{enumerate}
    \item \textbf{Initialization}. A population of policies $\Pi = \{ \pi_1, \pi_2, \dots, \pi_{n_\pi}\}$ is randomly initialized by sampling a set of parameters $\mathbf{\theta} \in \Theta $  from a parameter space $\Theta \subseteq \mathbb{R}^{n_\theta}
$ which represents the set of all possible parameter vectors. The weights in each neural network policy are initialized using He initialization, which is well-suited for networks with ReLU activations \citep{thimm1995neural}. This initialization method scales the variance of the weights to avoid issues like dead neurons and ensures stable gradient flow, promoting efficient learning in high-dimensional parameter spaces. Each policy $\pi_i$ is parameterized by $\mathbf{\theta}_i \in \mathbb{R}^{n_{\theta}}$ where $i \in \{1, 2, \cdots, n_\pi\}$ indexes the policies in the population and $n_\pi$ is the number of policies in the population. 
    \item \textbf{Policy Evaluation}. Each policy $\pi_i$ is evaluated in a MO-MDP environment. For each episode, the policy interacts with the environment where:
    \begin{itemize}
        \item At each time step $t$, given the current state $\mathbf{s}_t$, the policy select an action $\mathbf{a}_t$ where $\mathbf{a}_t \sim \pi_{\mathbf{\theta}_i} (\cdot | \mathbf{s}_t)$
        \item The environment transitions to the next state $\mathbf{s}_{t+1}$ given the MDP stochastic dynamics $P(\mathbf{s}_{t+1}|\mathbf{s}_t, \mathbf{a}_t)$ and receives a reward $r_t^j$ for each objective $j$ where $r_t^j = \mathcal{R}_j(\mathbf{s}_t,\mathbf{a}_t)$
        \item The policy accumulates the discounted rewards $R^j$ for each objective $j \in \{1, \cdots, n_f\}$, over the episode length $t_{tot}$ where $R^j(\pi_{\theta_i}) = \sum_{t=0}^{t_{tot}}\gamma^t r_t^j$
        \item This is repeated for $n_e$ episodes, and the average performance across episodes is calculated for each objective. $f_i^j = \frac{1}{n_e}\sum_{e=1}^{n_e} R^j(\pi_{\theta_i})$ where $\mathbf{f}_i = [f_i^1, \dots, f_i^{n_f}]$ represents the objective values for policy $\pi_{\theta_i}$.
    \end{itemize}
    \item \textbf{Non-Dominated Sorting}. Each solution is sorted into \textit{non-dominated} \textit{fronts} based on dominance relationships, as in NSGA-II. The first front $\mathcal{F}_1$ contains all solutions that are non-dominated with respect to the entire population. Subsequent fronts, denoted $\mathcal{F}_k$ for $k = 1,2,\ldots, K$ are then identified by recursively finding non-dominated solutions from the remaining populations. This process assigns a rank to each solution, with $\mathcal{F}_1$ representing the highest rank (best solution), $\mathcal{F}_2$ the second highest, and so on. 
    Each solution $\pi_i$ belongs to a front based on dominance where we identify dominance of one policy $\pi_i$ over another $\pi_l$ if:
    \begin{equation}
        \mathbf{f}_i \leq \mathbf{f}_l   \quad \text{and} \quad \mathbf{f}_i \neq \mathbf{f}_l
    \end{equation}
    This first conditions implies that that for all objectives $j = 1,2,\dots,n_f$; $f_i^j \leq f_l^j$ so policy $\pi_i$ performs at least as well as policy $\pi_l$ in all objectives. The second condition means that for at least one objective $j$, the performance of policy $\pi_i$ is strictly better than policy $\pi_l$ and hence guarantees that policy $\pi_i$ dominates $\pi_l$, i.e., $\pi_i \prec \pi_j$ where $\prec$ denotes the dominance relationship.
    \item \textbf{Crowding Distance Calculation}. The crowding distance $d(\pi_{\theta_i})$ is computed for each solution in the sorted fronts. 
    \item \textbf{Selection and Reproduction}. Binary tournament selection is applied, where policies are selected based on their non-dominance rank and crowding distance. The selected policies undergo crossover and mutation generating a set of offspring policies. 
    \begin{equation}
        \theta_\text{offspring} = \text{Crossover}(\theta_i, \theta_l) + \text{Mutation}(\theta_i')
    \end{equation}
    \item \textbf{Survival Selection}. The next generation $\Pi'$ of policies is formed by selecting the top $N$ policies based on their non-dominance rank and crowding distance from the combined population of parent population $\Pi$ and offspring $\Pi_\text{offspring}$. 
    \begin{equation}
        \Pi' = \text{Top-N}(\Pi \cup \Pi_\text{offspring})
    \end{equation}
    \item \textbf{Termination Criteria}. The process repeats until a termination criteria is met, i.e.,maximum generation number has been achieved or convergence criterion is met. 
    \item \textbf{Pareto Front}. The non-dominated solutions remaining in the final population represent the undominated Pareto front set of policies. 
    \begin{equation}
        \text{PF}(\Pi) = \left\{ \pi \in \Pi \mid \nexists \pi' \in \Pi : V_{\pi'} \succ_P V_{\pi} \right\}
    \end{equation}
    where $\succ_P$ is the Pareto dominance relation. This shows that every policy in the Pareto Front is non-dominated i.e., there is no other policy that performs better in every single objective. 
\end{enumerate}

\begin{algorithm}
\caption{MORSE}
\label{alg:morse}
\begin{algorithmic}[1]
\State \textbf{Input:} Number of policies $n_\pi$, Maximum generations $n_g$, Evaluation episodes $n_e$
\State \textbf{Output:} Pareto front set of policies $\mathcal{F}_{\text{Pareto}}$
\State \textbf{Step 1: Initialization}
\State Generate a population of policies $\Pi = \{ \pi_1, \pi_2, \ldots, \pi_{n_\pi}\}$, where each policy $\pi_i$ is parameterized by $\mathbf{\theta}_i$ \Comment{Initialize population}

\State \textbf{Step 2: Policy Evaluation}
\For{each policy $\pi_{\theta_i} \in \Pi$}
    \For{each episode $e = 1, 2, \ldots, n_e$}
        \State Initialize state $\mathbf{s}_0$ from the environment \Comment{Reset environment for new episode}
        \For{each time step $t = 0, 1, \ldots, t_{tot}$}
            \State Select action $\mathbf{a}_t \sim \pi_{\mathbf{\theta}_i}(\cdot|\mathbf{s}_t)$ based on the policy \Comment{Policy inference}
            \State Execute action $\mathbf{a}_t$, observe reward $\mathbf{r}_t$ and next state $\mathbf{s}_{t+1}$ \Comment{Environment interaction}
            \State Accumulate discounted reward for each objective $j$:
            \State $R^j(\pi_{\theta_i}) = \sum_{t=0}^{t_{tot}} \gamma^t r_t^j$ \Comment{Calculate episode reward}
        \EndFor
    \EndFor
    \State Compute average objective values over $n_e$ episodes for policy $\pi_{\theta_i}$:
    \State $\mathbf{f}_i = \frac{1}{n_e} \sum_{e=1}^{n_e} R^j(\pi_{\theta_i})$ \Comment{Average fitness over episodes}
\EndFor

\State \textbf{Step 3: Non-dominated Sorting}
\State Sort policies into fronts based on dominance relationships:
\State $\mathcal{F}_1, \mathcal{F}_2, \ldots \mathcal{F}_K$ \Comment{Apply non-dominated sorting algorithm}

\State \textbf{Step 4: Crowding Distance Calculation}
\For{each front $\mathcal{F}_k$}
    \State Compute crowding distance $d(\pi_{\theta_i})$ for each policy $\pi_{\theta_i}$:
    \State $d(\pi_{\theta_i}) = \sum_{j=1}^{n_f} \left( \frac{f_{j, \text{next}} - f_{j, \text{prev}}}{f_{j, \text{max}} - f_{j, \text{min}}} \right)$ \Comment{Calculate crowding distance for diversity}
\EndFor

\State \textbf{Step 5: Selection and Reproduction}
\State Perform binary tournament selection based on rank and crowding distance, followed by crossover and mutation:
\State $\theta_{\text{offspring}} = \text{Crossover}(\theta_{i}, \theta_{l}) + \text{Mutation}(\theta_{i}')$ \Comment{Generate new policies}

\State \textbf{Step 6: Survival Selection}
\State Combine parent population $\mathcal{P}$ and offspring $\mathcal{P}_{\text{offspring}}$, then select top $N$ policies based on non-domination rank and crowding distance:
\State $\mathcal{P}' = \text{Top-}N(\mathcal{P} \cup \mathcal{P}_{\text{offspring}})$ \Comment{Select next generation population}

\State \textbf{Step 7: Termination Criteria}
\If{Termination criteria met (e.g., $g \geq n_g$)}
    \State \textbf{break} \Comment{Exit loop if criteria satisfied}
\EndIf

\State \textbf{Step 8: Pareto Front Identification}
\State Identify non-dominated solutions in the final population:
\State $\mathcal{F}_{\text{Pareto}} = \{ \pi_i : \pi_i \text{ is non-dominated in } \mathcal{P}' \}$ \Comment{Extract final Pareto front}

\State \textbf{Return} Pareto front set $\mathcal{F}_{\text{Pareto}}, \quad \Pi^* = \{\pi_1^*, \dots, \pi_{n_{\pi}^*}^*\}$ \Comment{Return optimal policies}
\end{algorithmic}
\end{algorithm}

Our methodology offers a series of advantages compared to both traditional multi-objective optimization methods and other multi-objective reinforcement learning methods. Firstly, our methodology results in a Pareto front of policies rather than a single policy. This allows decision-makers to have the flexibility to switch between policies and choose one that gives the best trade-off according to the needs in real-time. This is valuable in dynamic environments where priorities may shift due to external disruptions to the system. Traditional MORL methods often reduce multiple objective into a single objective using scalarization. However, this requires prior knowledge of how to weight each objective. Our methodology avoids this as it optimizes for multiple objectives without requiring predefined weights. The non-dominated sorting and crowding distance mechanisms also ensure there is a good balance between exploring new solutions and exploiting known good policies, leading to a diverse set of policies across the Pareto front. By promoting a diverse set of solutions in the Pareto front, this enhances the chances of finding globally optimal solutions as evolutionary strategies perform a global search across the policy parameter space, reducing the likelihood of getting trapped in a local optima. Moreover, the population-based nature of evolutionary algorithms make them parallelizable as each policy can be evaluate independently. Additionally, because evolutionary algorithms do not rely on gradient-based optimization methods, they are more robust to noise and uncertainty in the environment which are key characteristics in inventory management systems. 
\section{Adaptive Behavior Analysis} \label{result}
In this section, we present the results obtained from applying our Multi-Objective Reinforcement Learning via Strategy Evolution (MORSE) framework to inventory management under two distinct scenarios: (1) emission penalties and (2) geopolitical tensions. We analyze three distinct supply chain configurations, summarized in Section \ref{sec:sc_configs}, to assess how the framework adapts to system shocks in varying structural setups. By leveraging our approach, we generate a Pareto set of policies that effectively balance competing objectives. Therefore, we analyze how our approach dynamically adapts to system disruptions, showcasing its ability to preserve profitability, mitigate environmental impacts, and enhance overall system resilience.

\subsection{Supply Chain Configurations} \label{sec:sc_configs}

To evaluate the robustness and adaptability of our Multi-Objective Reinforcement Learning via Strategy Evolution (MORSE) framework, we consider three distinct supply chain configurations. Each configuration varies in terms of network complexity and demand characteristics, enabling us to assess performance across a range of realistic operational environments.

\begin{itemize}
    \item \textbf{Configuration A} consists of a three-node network handling two product types, with demand following a seasonal distribution. The network includes one supplier, one distribution center, and one customer node. The seasonal demand introduces predictable variability, requiring the system to adapt to cyclical peaks and troughs in order volume.
    \item \textbf{Configuration B} maintains the same three-node, two-product structure but introduces stochasticity through a Poisson demand distribution. Unlike the seasonal demand pattern, this setup features random fluctuations in demand, simulating environments where customer orders arrive unpredictably and require more reactive decision-making.
    \item \textbf{Configuration 3} increases the network complexity to a five-node system while retaining the two-product setup. Demand in this configuration again follows a Poisson distribution. This setup tests the framework’s ability to adapt in more complex, distributed supply networks with cyclical demand pressures.
\end{itemize}

Together, these configurations allow us to ensure that our method generalizes effectively across different supply chain structures and demand environments, validating its robustness under varying forms of disruption.

\subsection{Adaptive Behavior}
A key strength of our MORSE framework lies in its adaptive behavior under stochastic dynamically changing conditions. By generating a Pareto-optimal set of policies, the system maintains the flexibility to switch between policies in real time, depending on the preferred operator objectives and external constraints. This adaptability ensures that the system remains robust and responsive to diverse operational challenges, offering decision-makers a range of viable solutions for each scenario.

When disruptions such as emission penalties or geopolitical cost fluctuations occur, the Pareto set allows for rapid policy adjustments. As these disruptions alter the operational landscape, the system can dynamically select the most suitable policy to balance trade-offs between competing objectives, such as minimizing costs, reducing emissions, and ensuring service level.

\subsubsection{Emission Penalties}
As environmental concerns continue to grow, governments are increasingly implementing regulations to mitigate the impacts of climate change. These regulations, such as emission taxes, not only directly affect profitability but also require firms to adapt their operational strategies in order to meet environmental standards. In this case study, we simulate the impact of an emission tax, which penalizes firms when emissions exceed a predefined threshold within a given time period. To evaluate the system’s adaptability, we introduce the emission tax at time step 200, and the policy is switched to a more favourable one on the Pareto front. The objective is to evaluate how the MORL framework allows the operator to switch to policies that favour reducing emissions to protect profits when the emission tax is reduced. We compare the overall system performance between dynamically switching policies and operating at the same policy level. 

\begin{figure}[htbp]
    \begin{subfigure}[b]{\textwidth}
        \centering
        \includegraphics[width=0.6\textwidth]{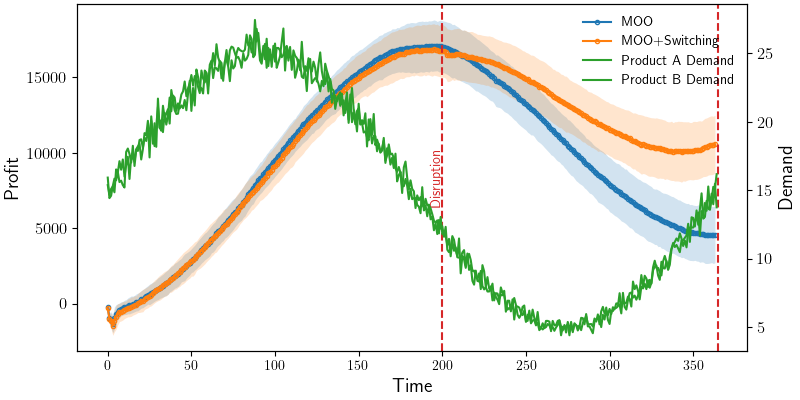}
    \end{subfigure}
    \vspace{0.5cm}
    \begin{subfigure}[b]{\textwidth}
        \centering
        \includegraphics[width=0.6\textwidth]{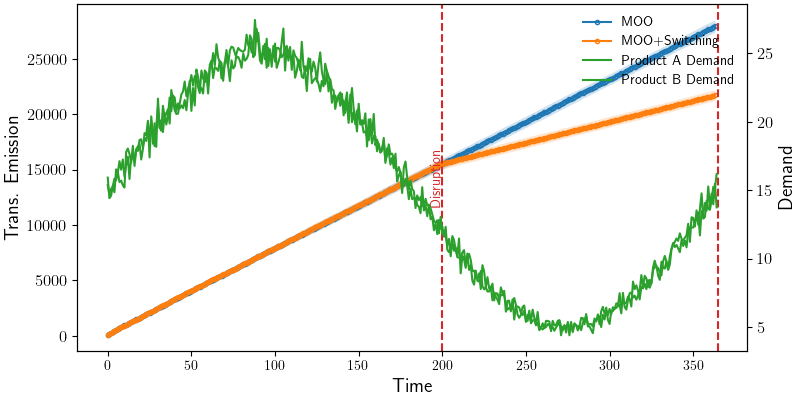}
    \end{subfigure}
    \vspace{0.5cm}
    \begin{subfigure}[b]{\textwidth}
        \centering
        \includegraphics[width=0.6\textwidth]{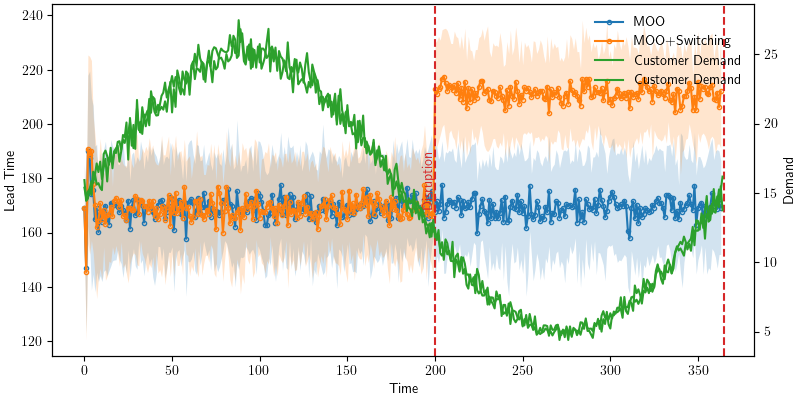}
    \end{subfigure}
    \caption{Dynamics of cumulative profit, cumulative transportation emission, and non-cumulative lead time under emission penalties scenario for Configuration A.}
    \label{fig:profit}
\end{figure}

\paragraph{Configuration A} As seen in Figure \ref{fig:profit}, the emission tax leads to a decline in cumulative profits due to penalties for exceeding the emission threshold. Our adaptive strategy protects profits and minimizes emissions to comply with environmental regulations. By adjusting operational strategies in real time, the system can balance the trade-off between profitability and sustainability, at the expense of higher lead times.

\paragraph{Configuration B}

\begin{figure}[htbp]
    \begin{subfigure}[b]{\textwidth}
        \centering
        \includegraphics[width=0.6\textwidth]{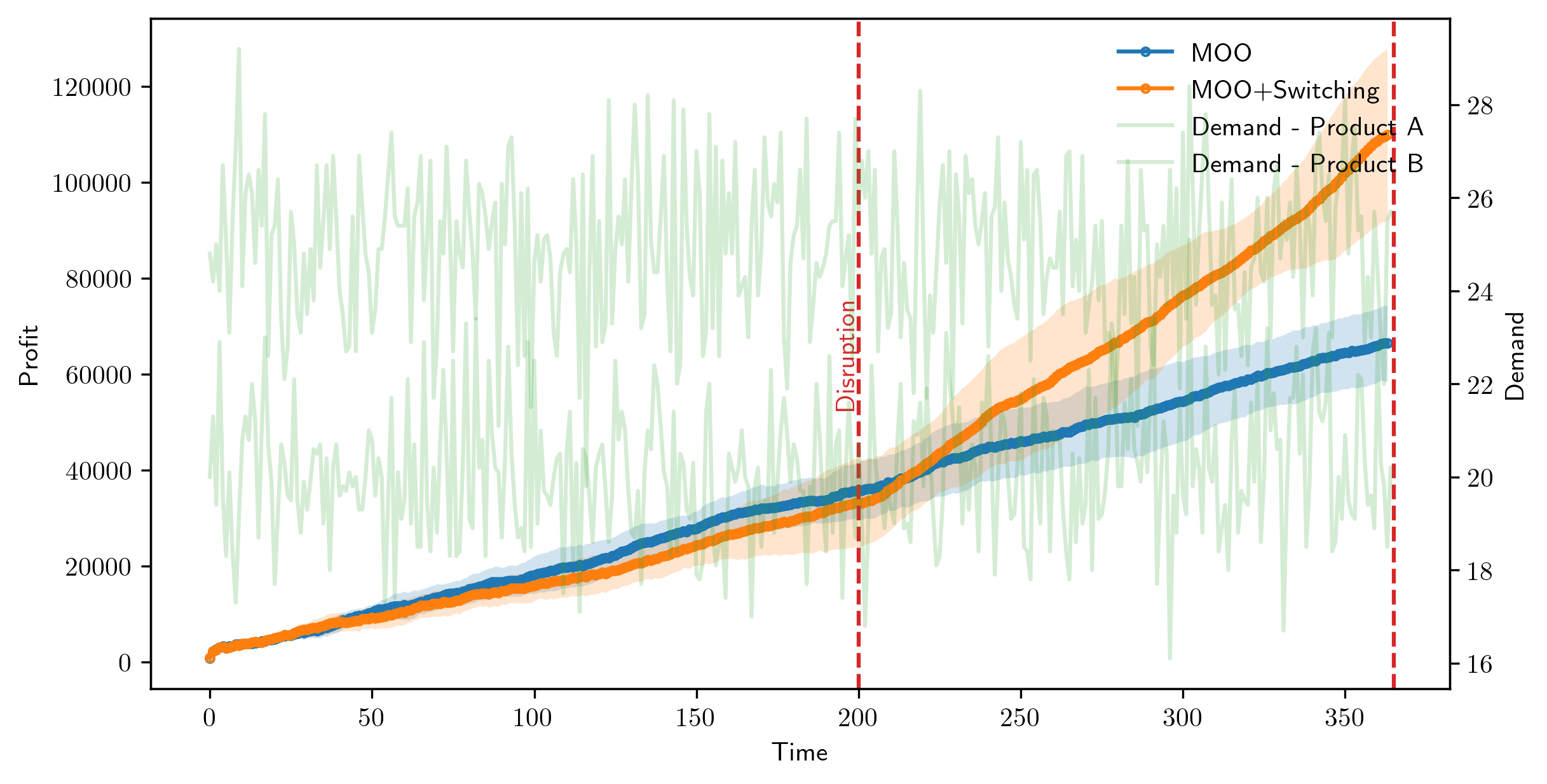}
    \end{subfigure}
    \vspace{0.5cm}
    \begin{subfigure}[b]{\textwidth}
        \centering
        \includegraphics[width=0.6\textwidth]{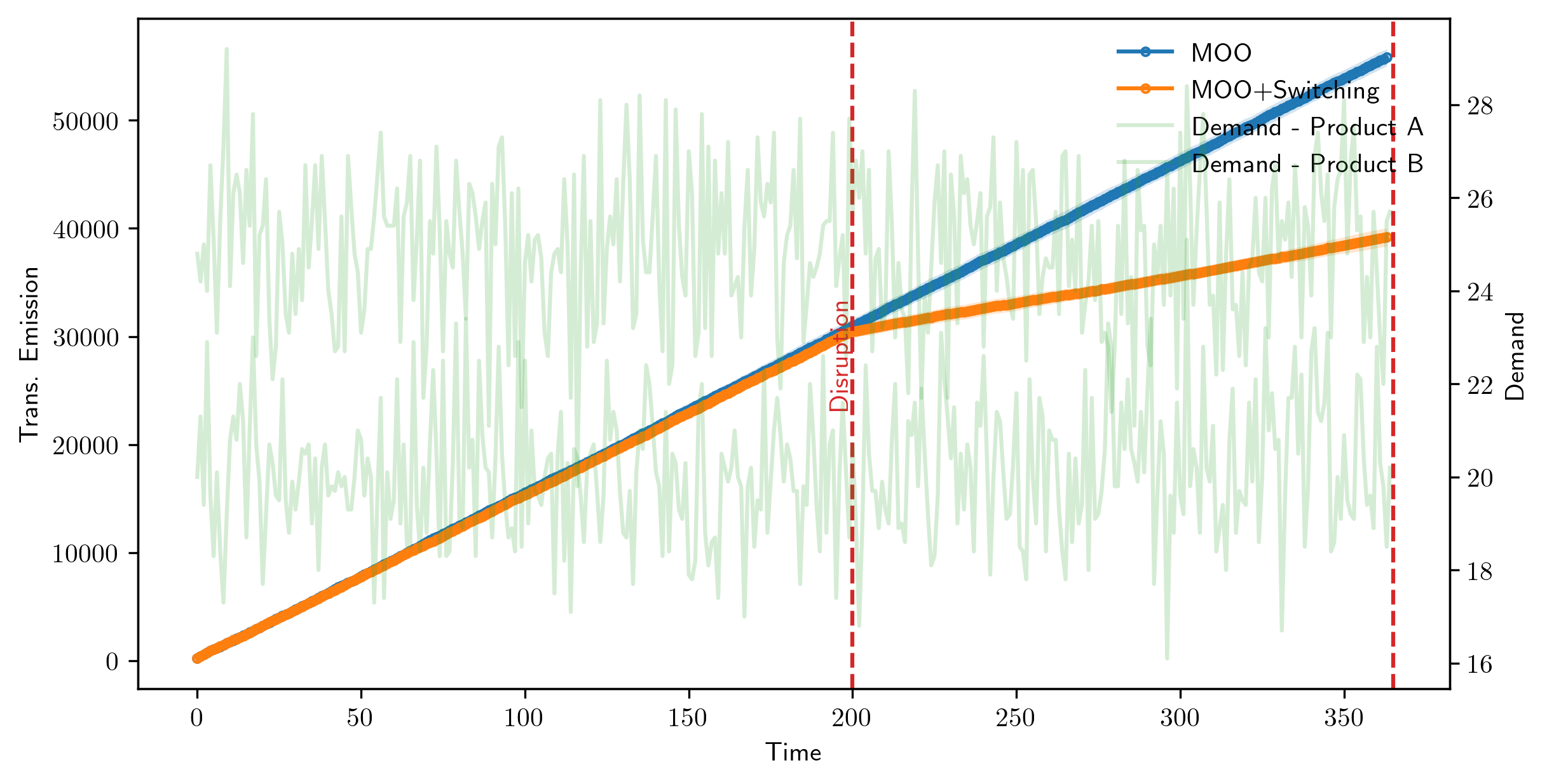}
    \end{subfigure}
    \vspace{0.5cm}
    \begin{subfigure}[b]{\textwidth}
        \centering
        \includegraphics[width=0.6\textwidth]{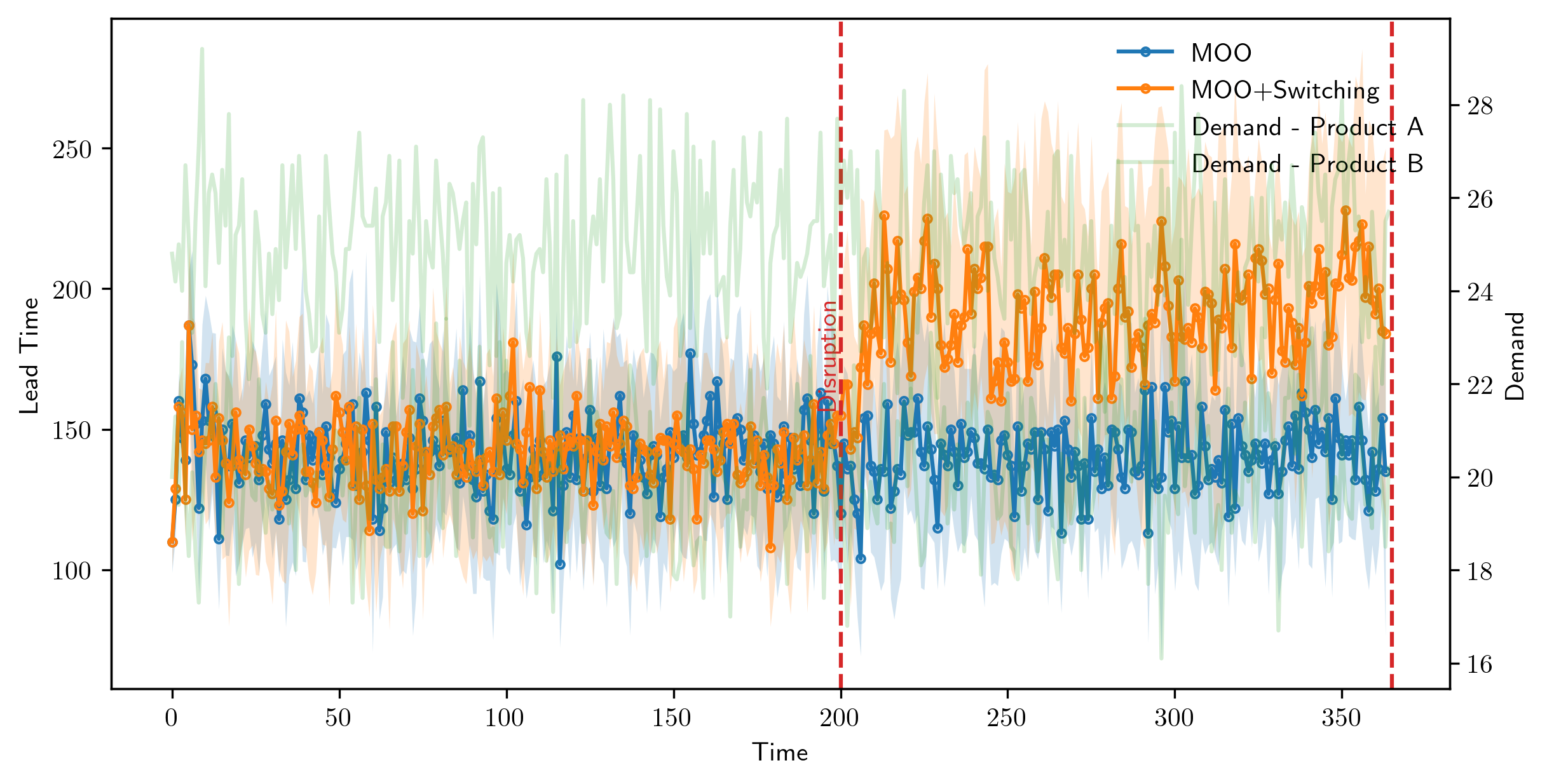}
    \end{subfigure}
    \caption{Dynamics of cumulative profit, cumulative transportation emission, and non-cumulative lead time under emission penalties scenario for Configuration B.}
    \label{fig:profit_B}
\end{figure}

\begin{figure}[htbp]
    \begin{subfigure}[b]{\textwidth}
        \centering
        \includegraphics[width=0.6\textwidth]{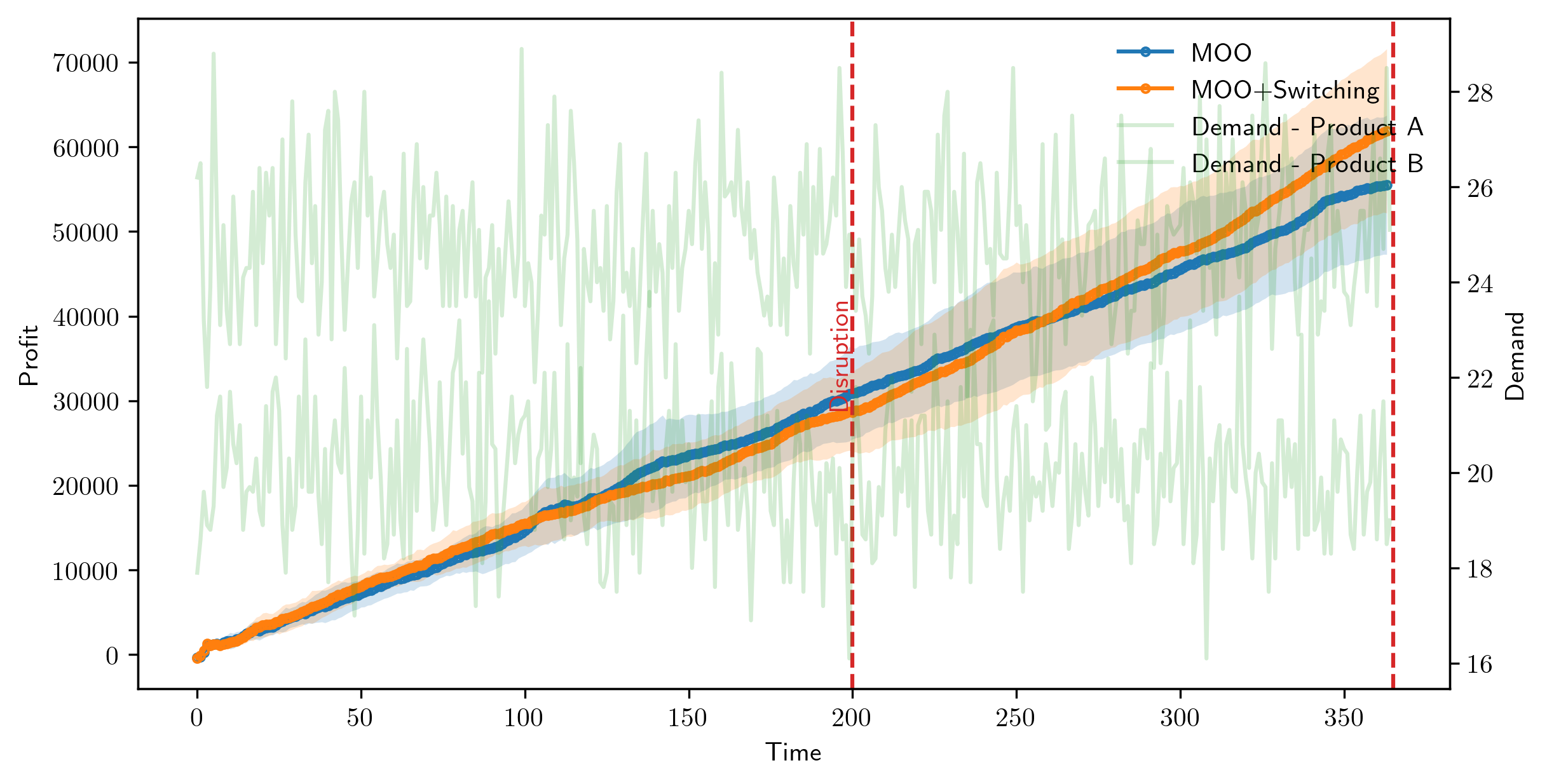}
    \end{subfigure}
    \vspace{0.5cm}
    \begin{subfigure}[b]{\textwidth}
        \centering
        \includegraphics[width=0.6\textwidth]{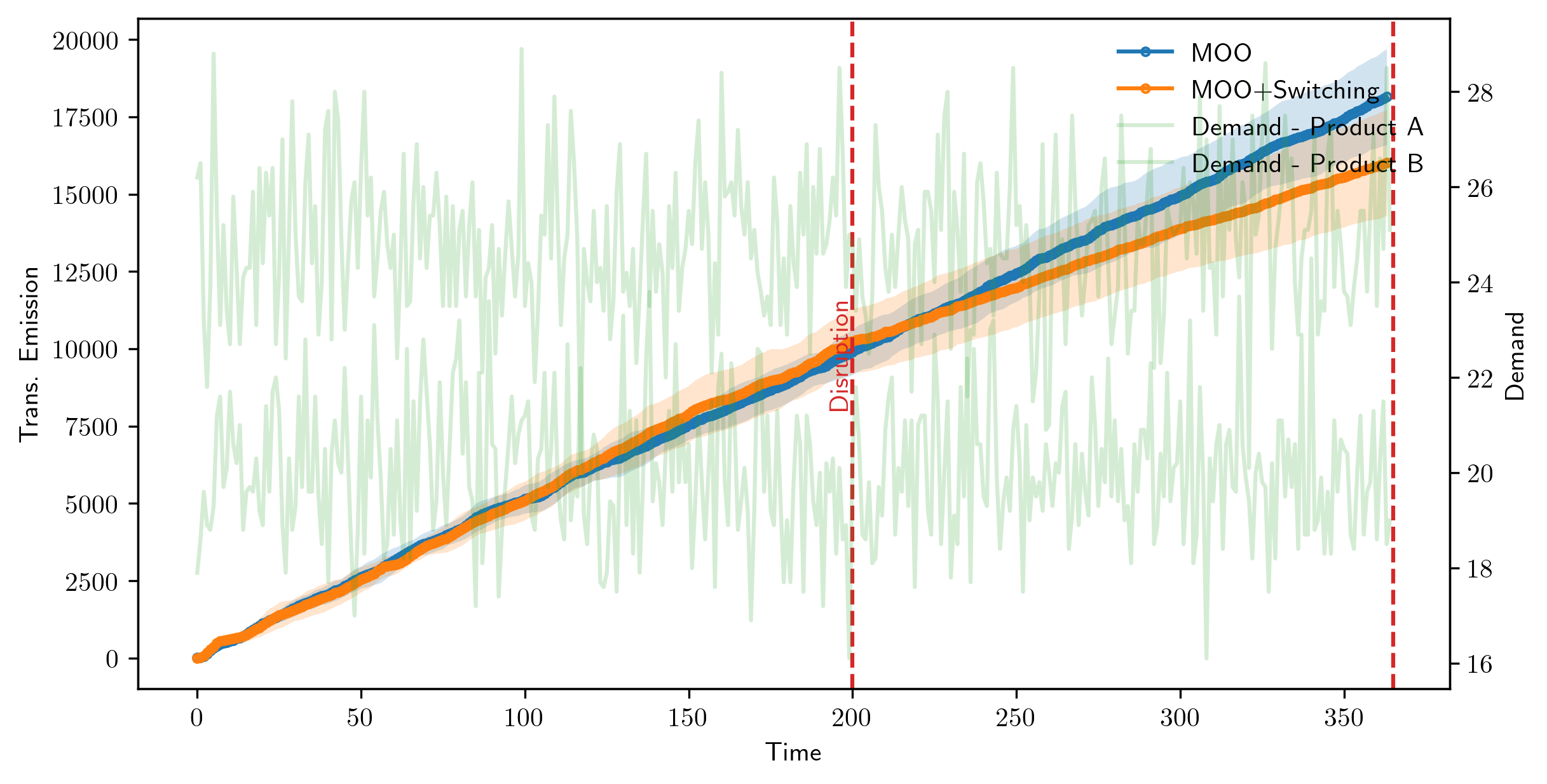}
    \end{subfigure}
    \vspace{0.5cm}
    \begin{subfigure}[b]{\textwidth}
        \centering
        \includegraphics[width=0.6\textwidth]{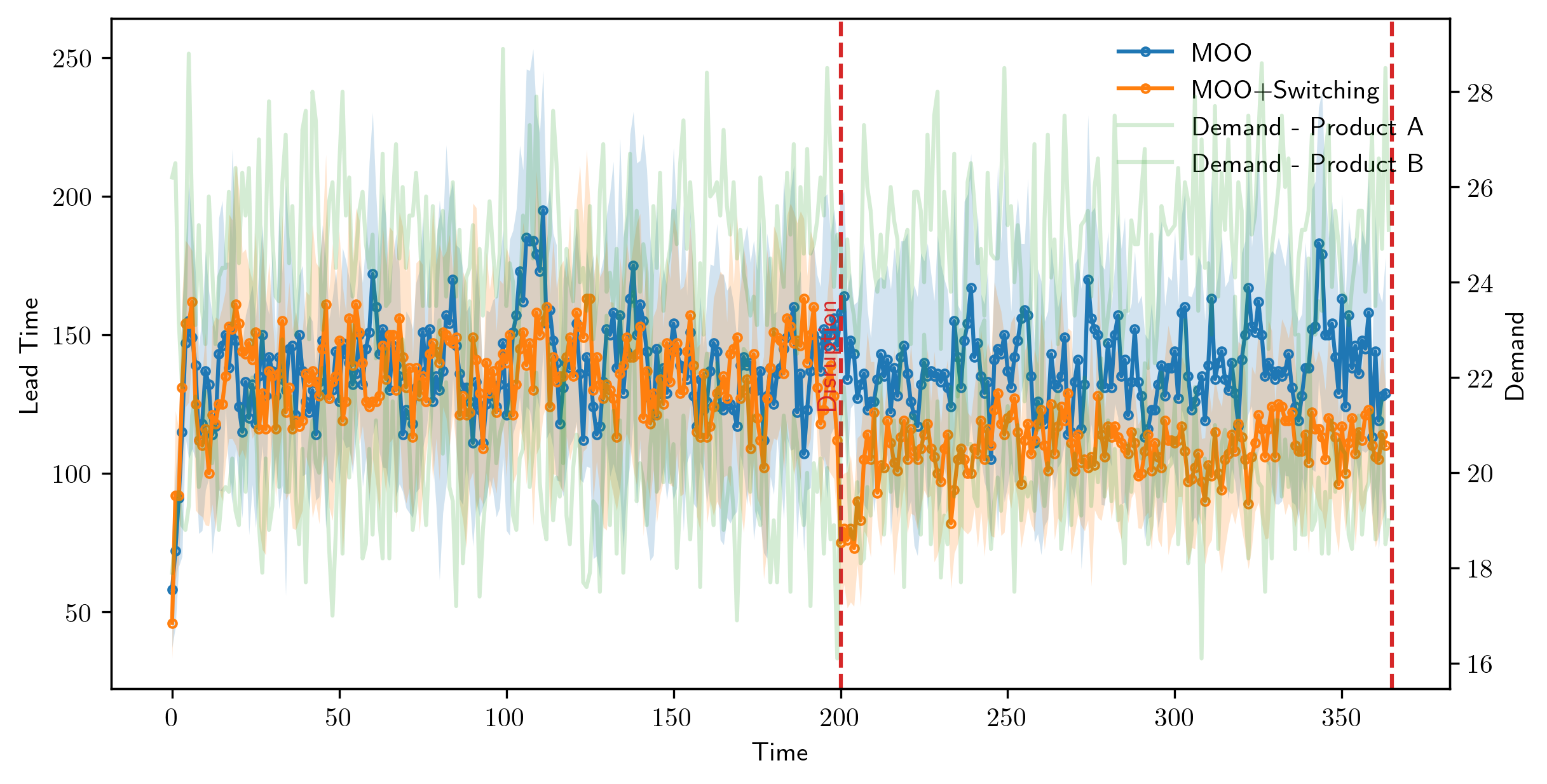}
    \end{subfigure}
    \caption{Dynamics of cumulative profit, cumulative transportation emission, and non-cumulative lead time under emission penalties scenario for Configuration C.}
    \label{fig:profit_C}
\end{figure}

\paragraph{Configuration C}
With a more complex network, the system demonstrates its ability to dynamically shift operational policies to mitigate emission penalties, sustaining profitability and controlling lead times despite the regulatory pressure as shown in Figure \ref{fig:profit_C}. 

 Therefore, by using our multi-objective approach, which results in a Pareto set of policies, the system is capable of making real-time adjustments and decisions based on the current state of the environment. This flexibility allows the system to effectively navigate trade-offs between competing objectives, ensuring that operational goals are met while adhering to system disruptions and constraints. As a result, the system exhibits enhanced resilience and adaptability, enabling it to respond dynamically to evolving conditions.
\subsubsection{Geopoltical Tensions}
In light of increasing geopolitical tensions, businesses are facing heightened risks and uncertainties that can lead to a rise in operational costs. Geopolitical disruptions, such as trade restrictions, sanctions, or supply chain instability, often require rapid strategic adjustments to maintain resilience and profitability. These conditions can lead to increased costs in procurement and transportation, directly impacting a firm’s financial performance. In this scenario, we simulate the impact of geopolitical tensions by increasing the costs by 10\% over a certain duration. This cost increase simulates the heightened financial pressure experienced by firms operating in volatile geopolitical environments. Therefore, from our trained set of policies, we require to switch to a policy that can mitigate the negative financial consequences of these cost surges while ensuring consistent service levels.

\begin{figure}[htbp]
    \begin{subfigure}[b]{\textwidth}
        \centering
        \includegraphics[width=0.6\textwidth]{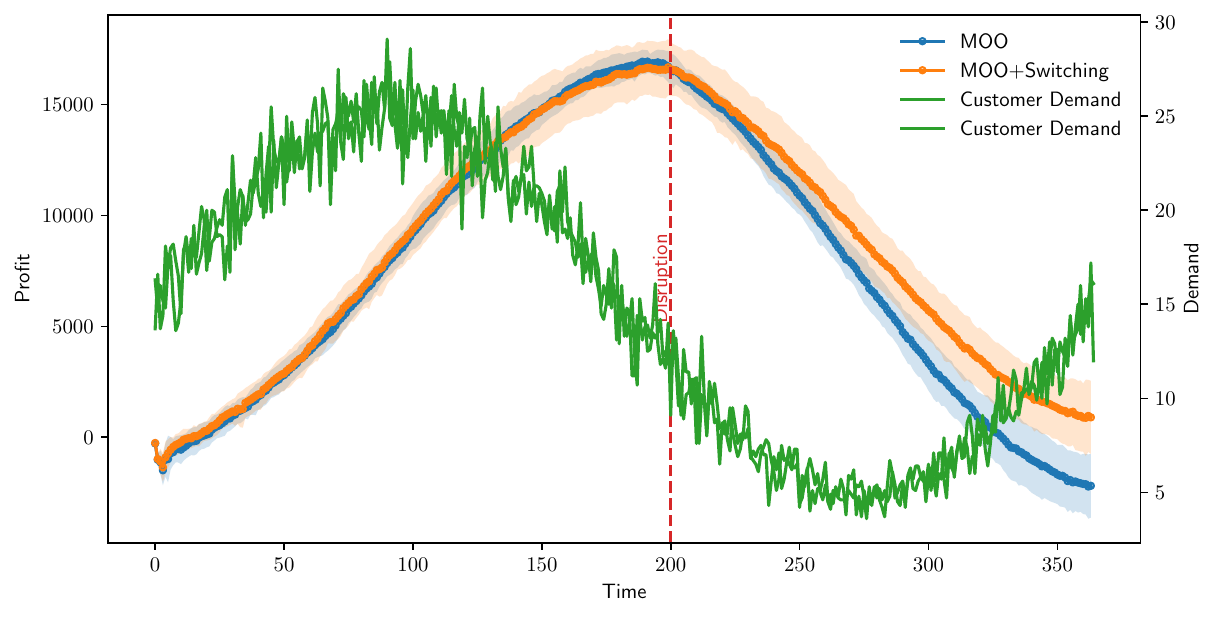}
    \end{subfigure}
    \vspace{0.5cm}
    \begin{subfigure}[b]{\textwidth}
        \centering
        \includegraphics[width=0.6\textwidth]{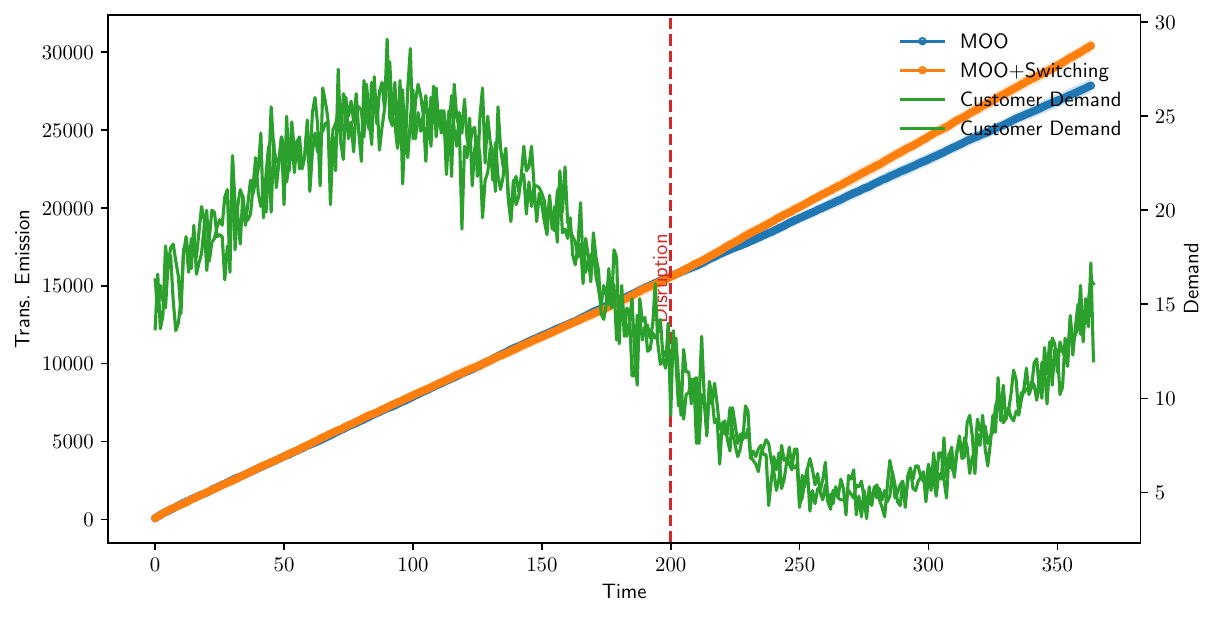}
    \end{subfigure}
    \vspace{0.5cm}
    \begin{subfigure}[b]{\textwidth}
        \centering
        \includegraphics[width=0.6\textwidth]{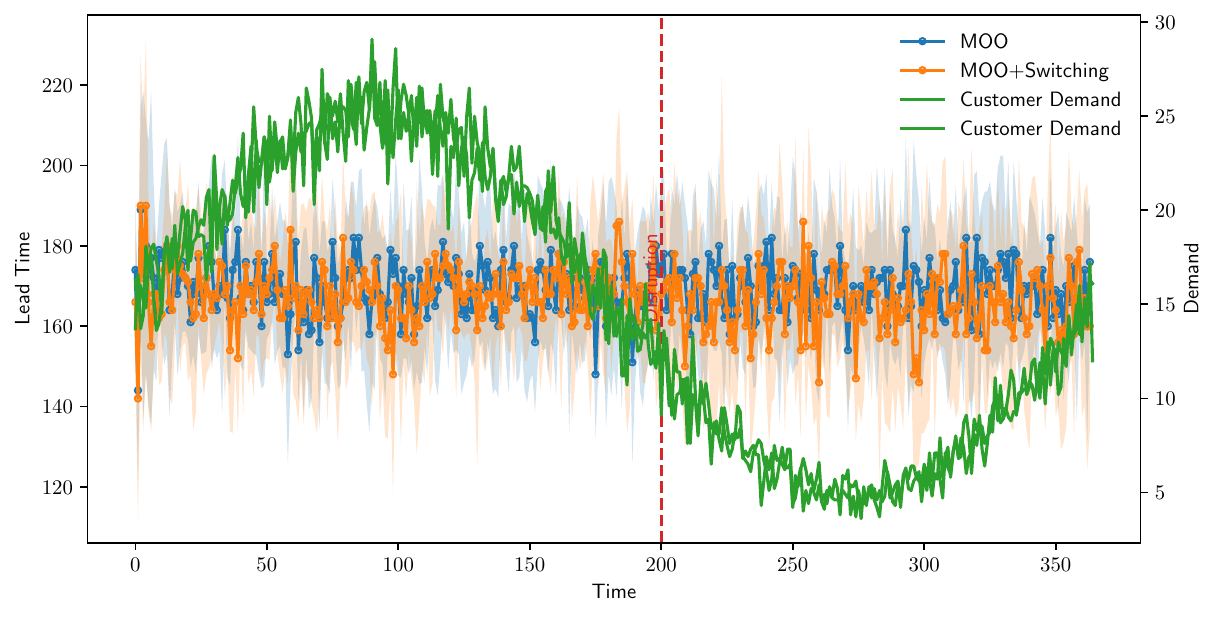}
    \end{subfigure}
    \caption{Dynamics of cumulative profit, cumulative transportation emission, and non-cumulative lead time under geopolitical tension scenario for Configuration A.}
    \label{fig:geo_A}
\end{figure}

\paragraph{Configuration A} As seen in Figure \ref{fig:geo_A}, the increase in costs due to geopolitical tensions leads to a decrease in cumulative profits. In response to these rising costs, our adaptive strategy aims to shield profits by dynamically adjusting operations. This ensures that the system can mitigate the impact of ongoing disruptions, optimizing performance despite the external challenges. By incorporating flexibility and real-time decision-making, our strategy allows for effective navigation through uncertain geopolitical environments, safeguarding long-term profitability.

\begin{figure}[htbp]
    \begin{subfigure}[b]{\textwidth}
        \centering
        \includegraphics[width=0.6\textwidth]{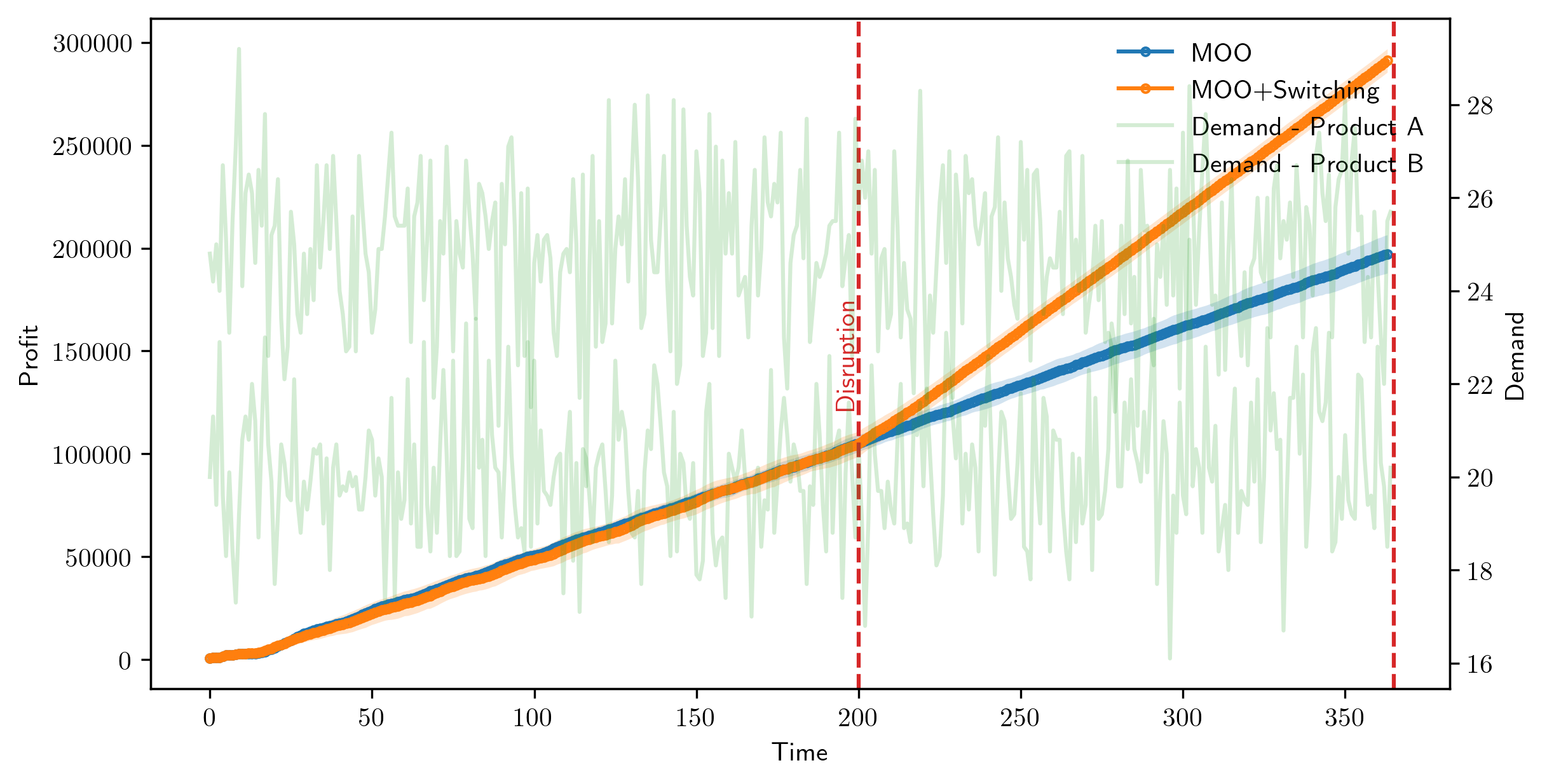}
    \end{subfigure}
    \vspace{0.5cm}
    \begin{subfigure}[b]{\textwidth}
        \centering
        \includegraphics[width=0.6\textwidth]{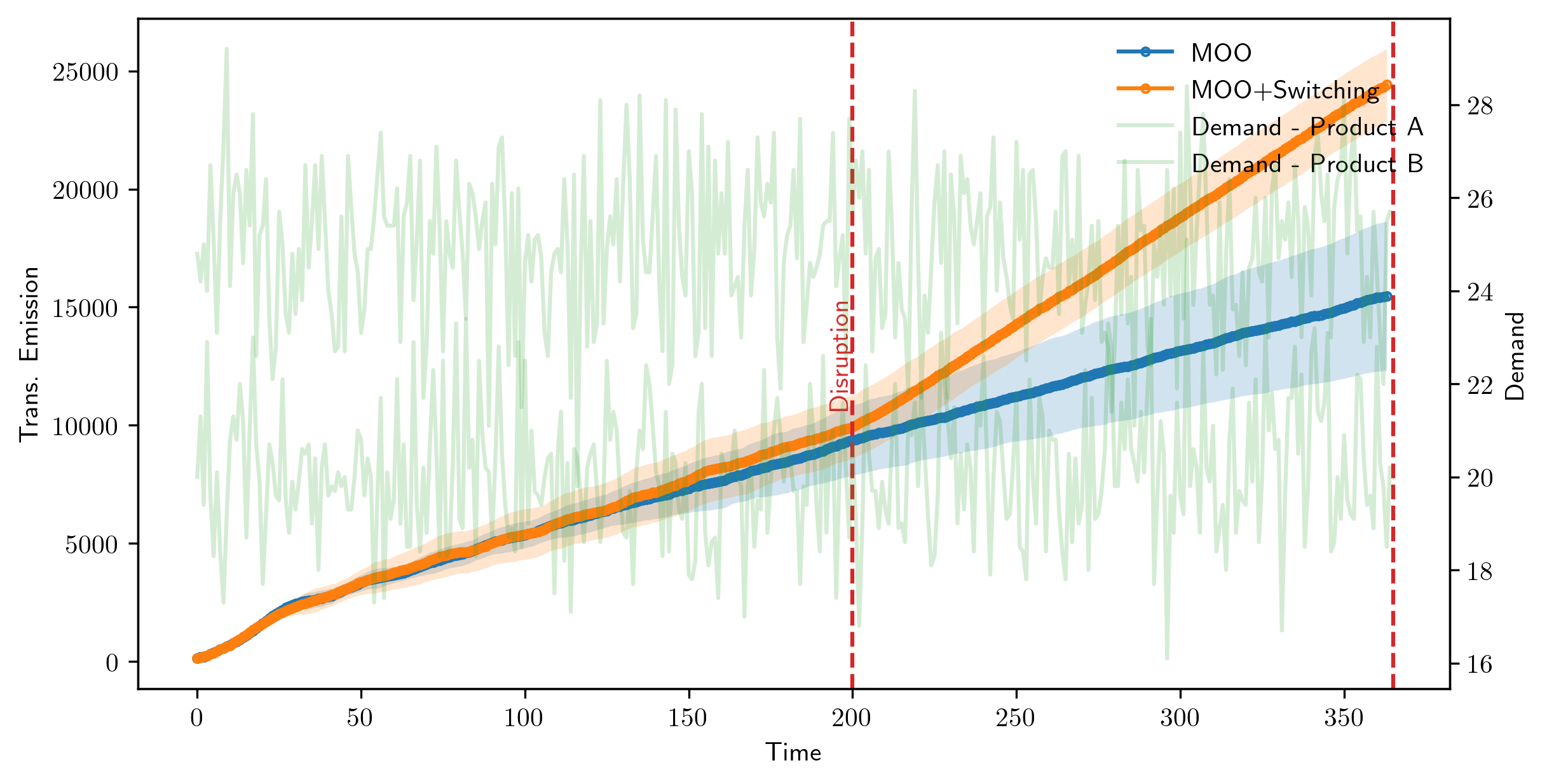}
    \end{subfigure}
    \vspace{0.5cm}
    \begin{subfigure}[b]{\textwidth}
        \centering
        \includegraphics[width=0.6\textwidth]{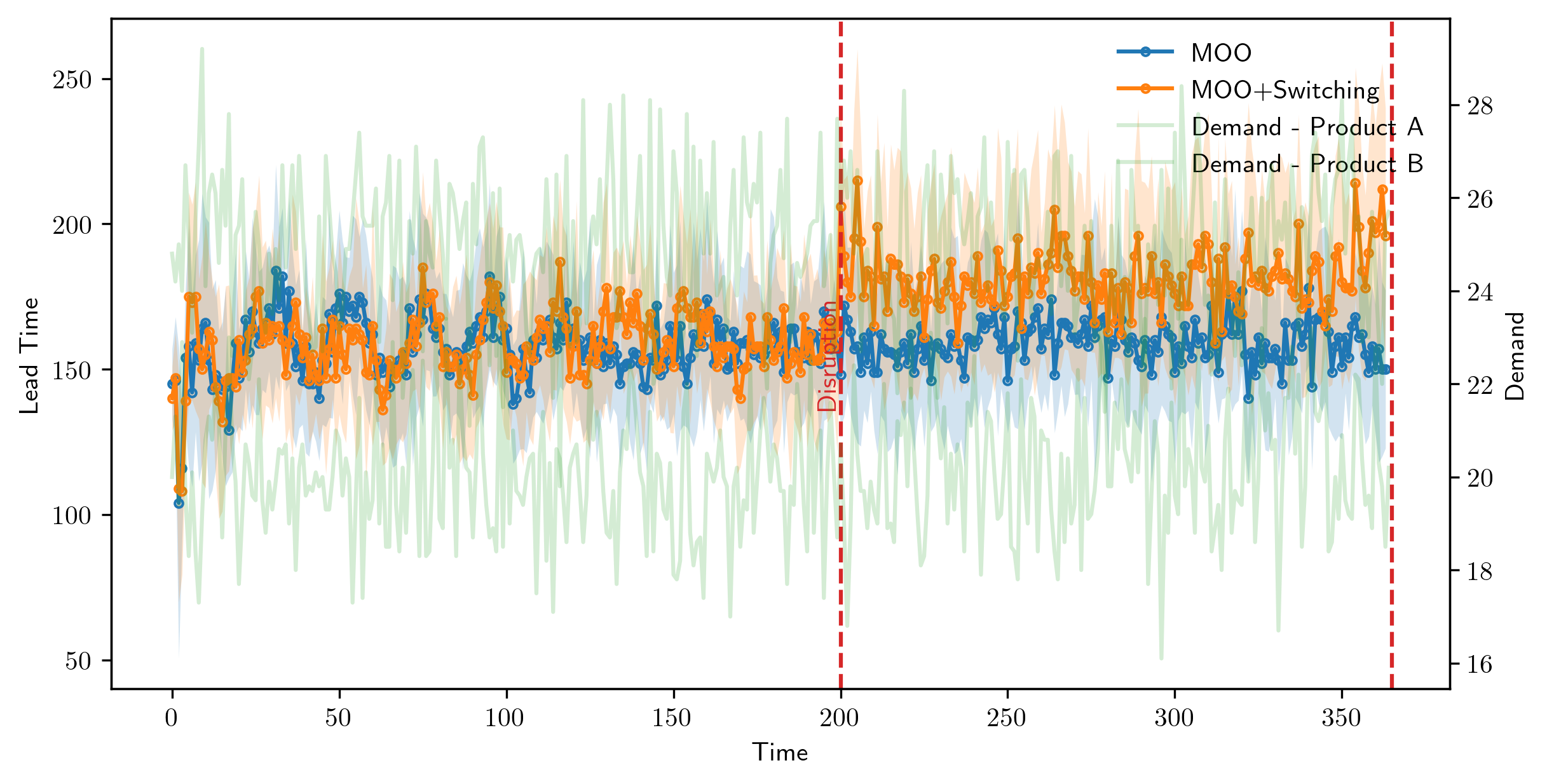}
    \end{subfigure}
    \caption{Dynamics of cumulative profit, cumulative transportation emission, and non-cumulative lead time under geopolitical tension scenario for Configuration B.}
    \label{fig:geo_B}
\end{figure}

\paragraph{Configuration B}
In this supply chain configuration, the system achieved higher profitability at the expense of higher emissions as shown in Figure \ref{fig:geo_B}.

\begin{figure}[htbp]
    \begin{subfigure}[b]{\textwidth}
        \centering
        \includegraphics[width=0.6\textwidth]{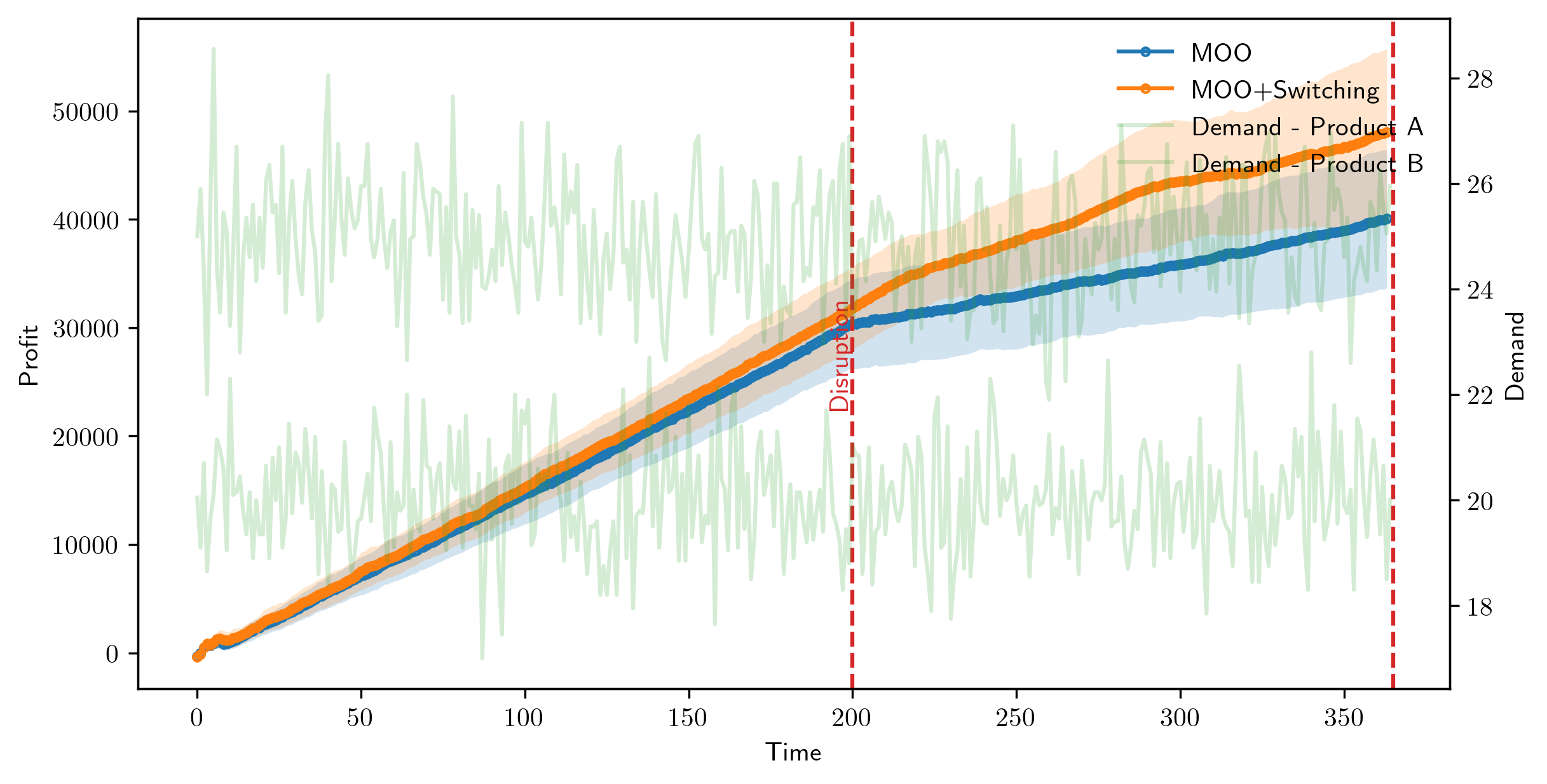}
    \end{subfigure}
    \vspace{0.5cm}
    \begin{subfigure}[b]{\textwidth}
        \centering
        \includegraphics[width=0.6\textwidth]{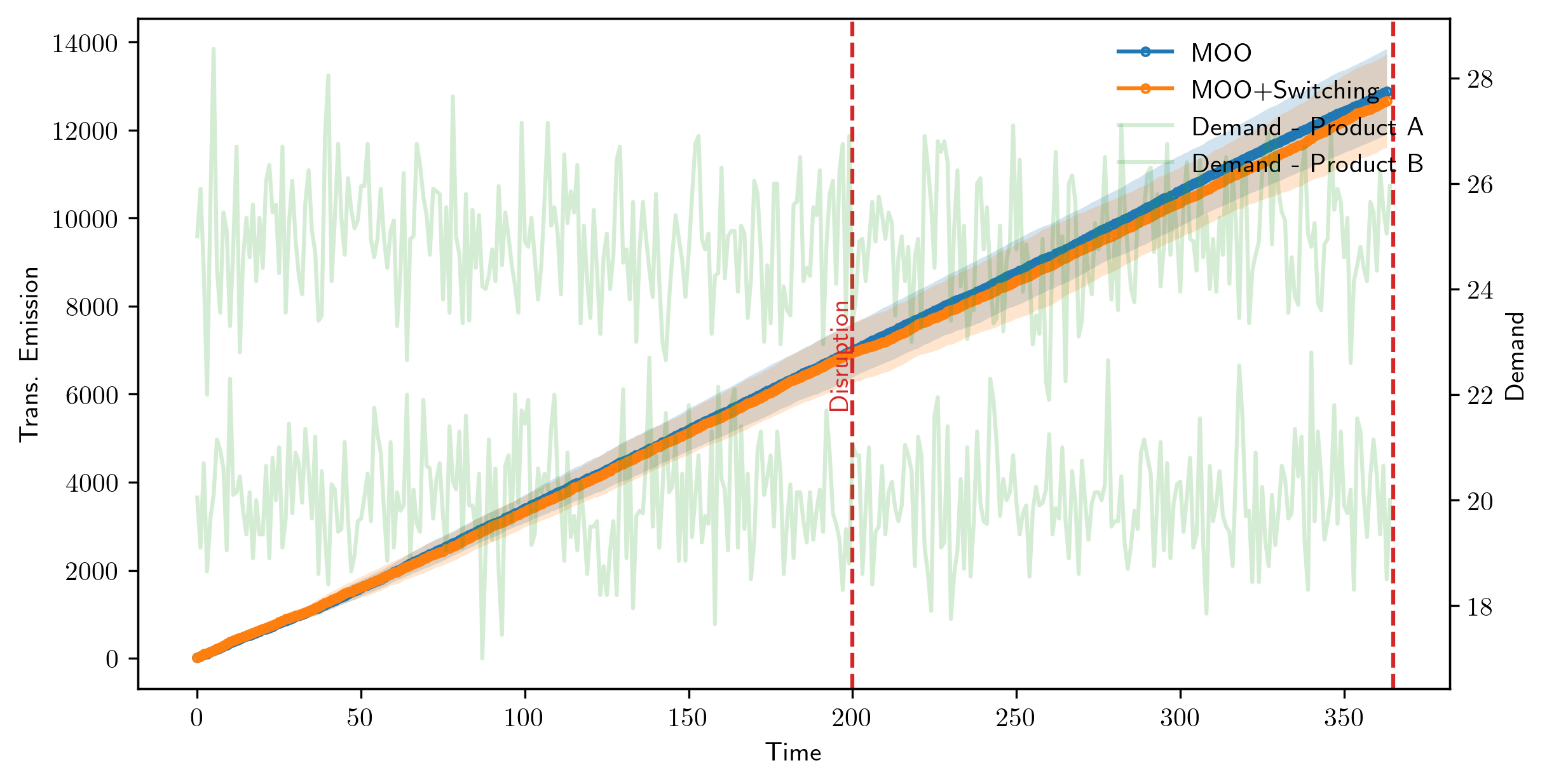}
    \end{subfigure}
    \vspace{0.5cm}
    \begin{subfigure}[b]{\textwidth}
        \centering
        \includegraphics[width=0.6\textwidth]{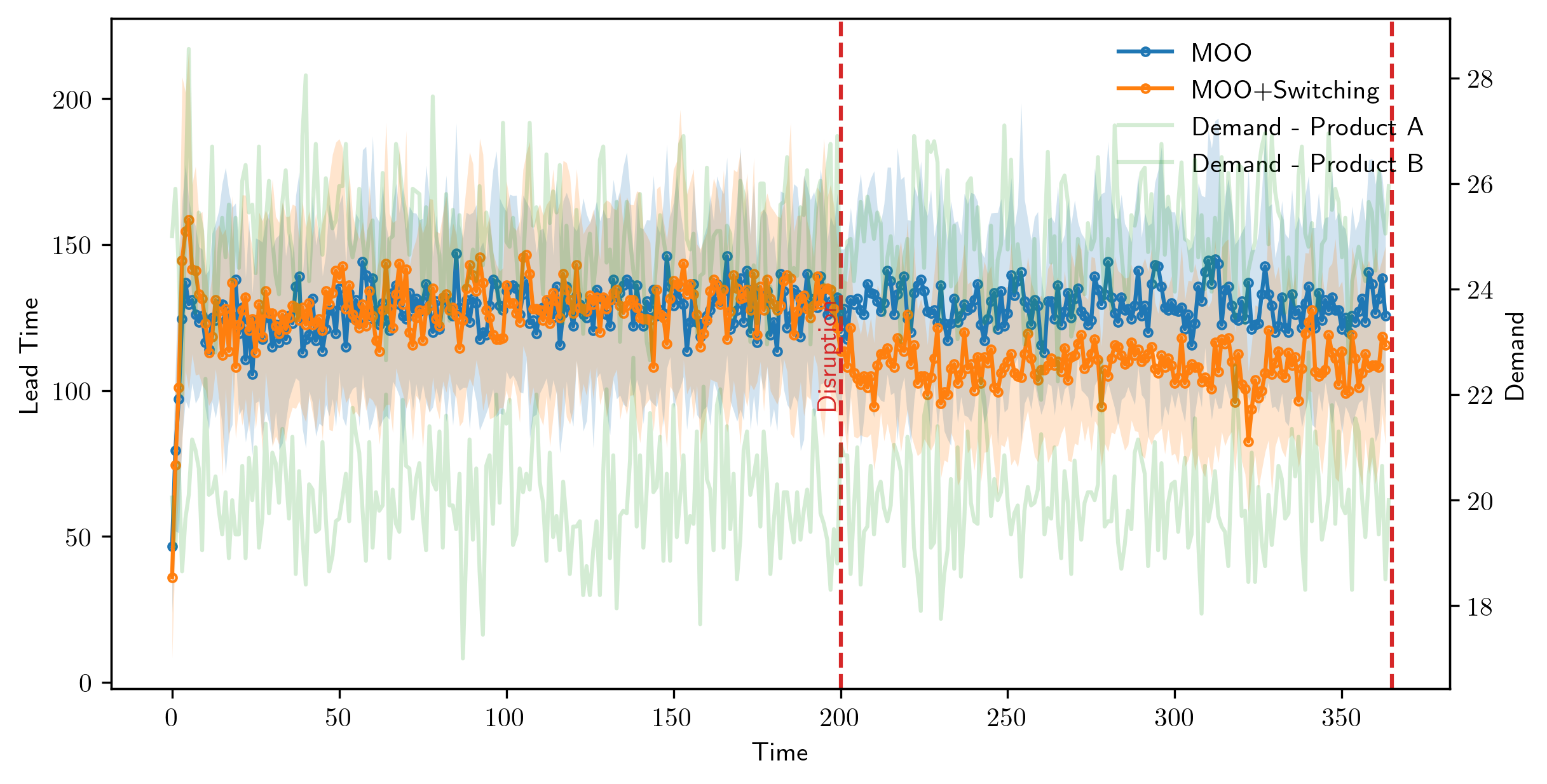}
    \end{subfigure}
    \caption{Dynamics of cumulative profit, cumulative transportation emission, and non-cumulative lead time under geopolitical tension scenario for Configuration C.}
    \label{fig:geo_C}
\end{figure}

\paragraph{Configuration C}
With a more complex network structure, this configuration shows how the framework dynamically navigates increased costs by optimizing between profitability and sustainability metrics under more diverse operational conditions as shown in Figure \ref{fig:geo_C}. 

Overall, across all three supply chain configurations, our results demonstrate that the MORSE framework successfully adapts to dynamic operational conditions in inventory management under both environmental and geopolitical disruptions. By generating a Pareto set of policies, the system provides decision-makers with a wide range of choices that balance profitability, resilience, and sustainability. The ability to switch between policies in real time enhances the system’s capacity for fast decision-making in the face of disruptions which is particularly important when navigating complex, uncertain environments. 

\section{Robustness Analysis}\label{robust}
When developing policies for safety-critical or operationally sensitive applications, it is essential to account for risks and prevent catastrophic outcomes. Therefore, simply focusing on maximizing the expected return does not safeguard the decision-maker against rare but high-impact events. To address this, in this section we analyze the impact of incorporating Conditional Value at Risk (CVaR) into the retraining process of the Pareto fronts to enhance policy robustness. CVaR, also known as the expected shortfall, provides a more comprehensive measure of tail risk by capturing the expected loss beyond a specified quantile threshold. This ensures the policies become more robust to extreme outcomes, ensuring that the solutions perform well under both typical and adverse conditions.

\subsection{Incorporating CVaR for Robust Pareto Fronts}
To improve the robustness of the Pareto solutions, we leverage the empirical distribution of returns derived from a number of sampled episodes and introduce the concepts of value at risk (VaR) and conditional value at risk (CVaR) to optimize the supply chain by accounting for tail risk in the return distribution. 

VaR, $V_\alpha$, is defined as the maximum possible return under a given policy, $\pi$. For a confidence level $\alpha \in (0,1)$:
\begin{equation}
V_\alpha = \max \{g : F_\pi(g) \leq 1 - \alpha\} \iff V_\alpha = F_\pi^{-1}(1 - \alpha)
\end{equation}
where $G_0$ is a random variable of return and $F_\pi(g) = \mathbb{P}(G_0 \leq g)$ is the cumulative distribution function of the performance associated with a policy \citep{rockafellar2000optimization}.

While value at risk (VaR) is a useful measure for assesing the maximum possible loss at a given confidence level, VaR only captures the threshold below which returns fall with a probability of $1-\alpha$, providing no information about the severity of losses beyond this threshold. This makes VaR insufficient for fully understanding the impact of extreme negative events. 

Therefore we use conditional value at risk (CVaR), which considers the expected return in the worst $\alpha-$tail of the distribution. 

\begin{figure}[htbp]
    \centering
    \begin{subfigure}[b]{0.45\textwidth}
        \centering
        \includegraphics[width=\textwidth]{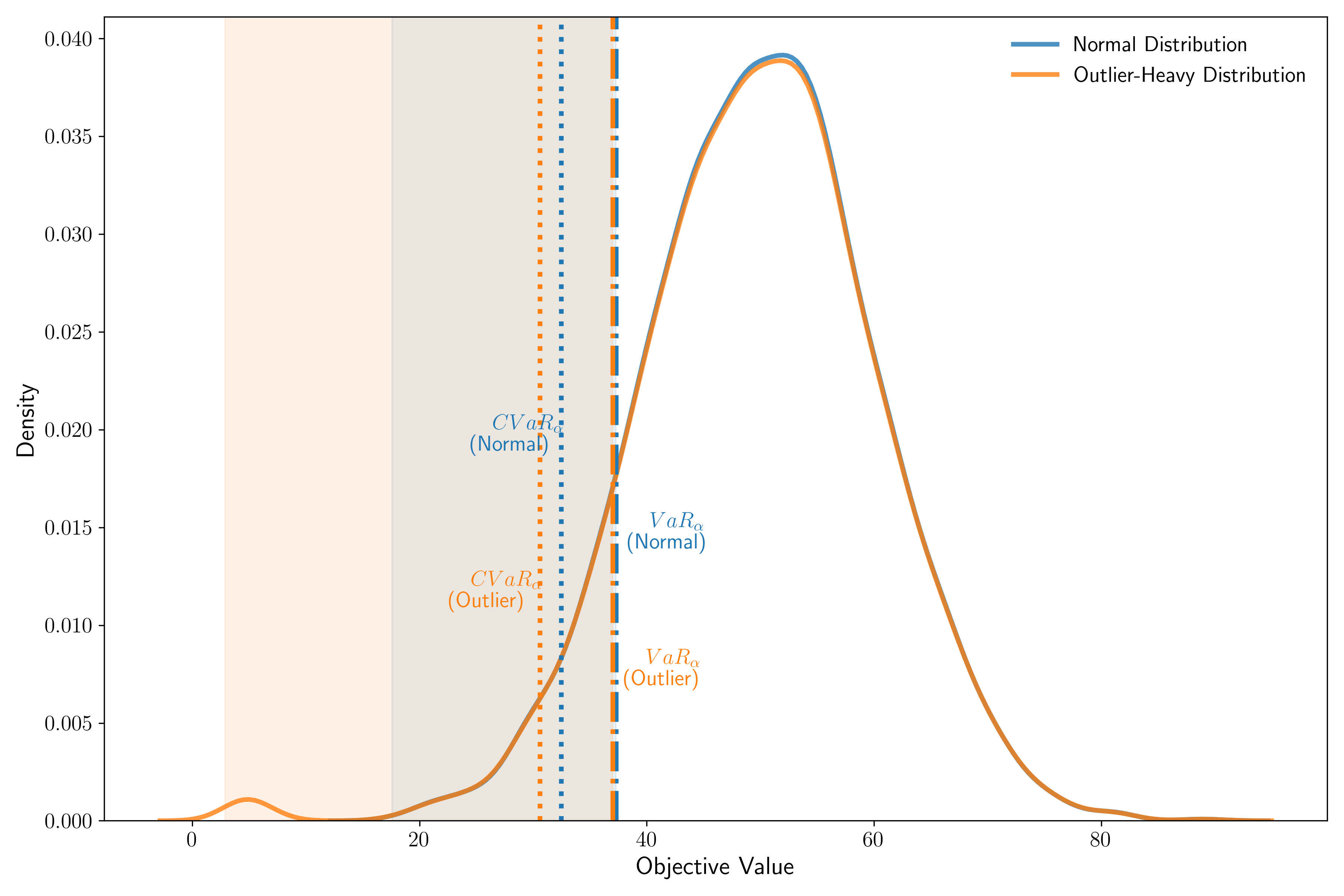}
        \caption{Density Plot.}
        \label{fig:histogram}
    \end{subfigure}
    \hfill
    \begin{subfigure}[b]{0.45\textwidth}
        \centering
        \includegraphics[width=\textwidth]{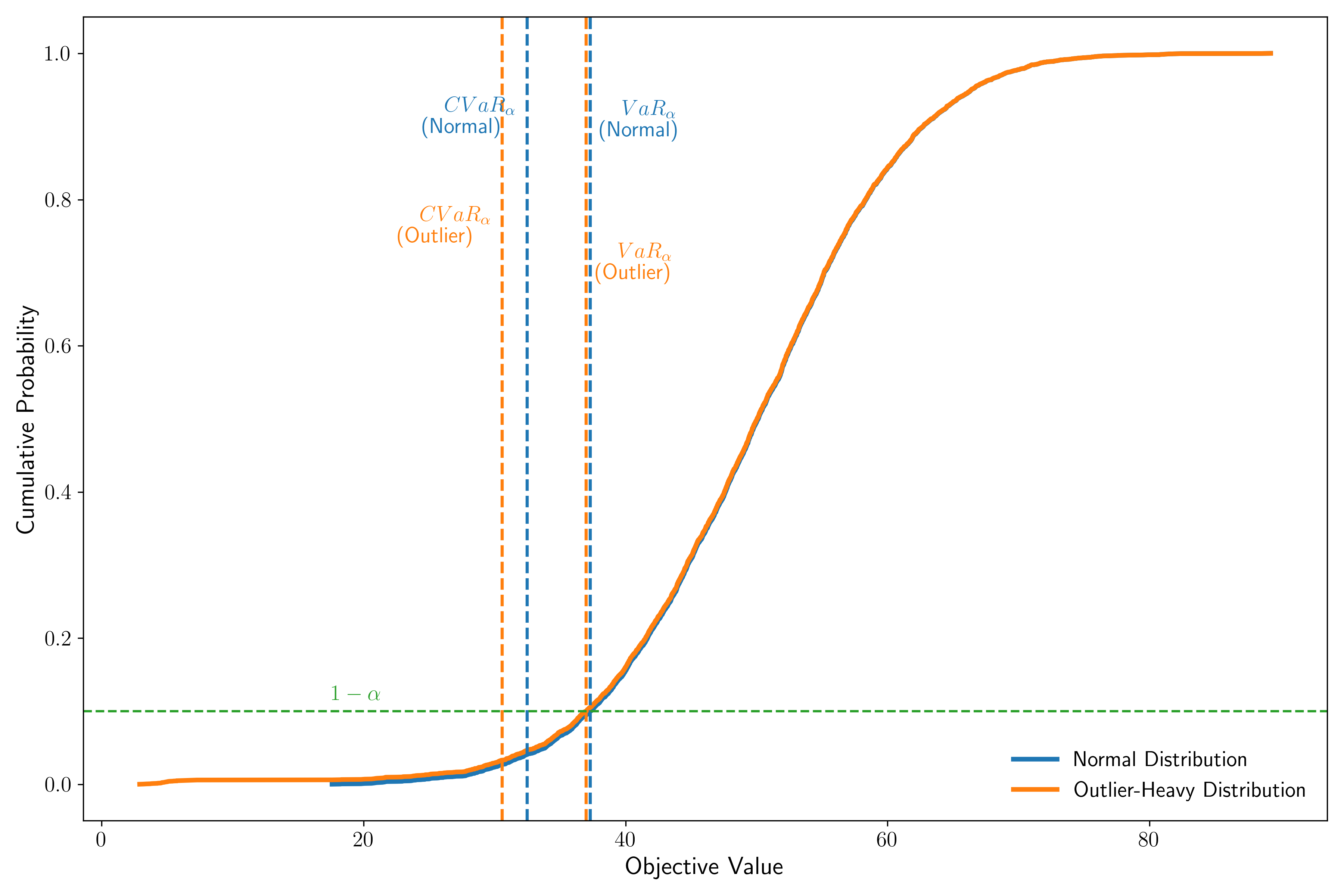}
        \caption{Cumulative distribution function (CDF)}
        \label{fig:cdf}
    \end{subfigure}
    \caption{Comparison of 1000 samples from a normal distribution with mean 50 and a heavy-tailed distribution with outliers. The histogram and CDF display the Value at Risk (VaR), and Conditional Value at Risk (CVaR) when $\alpha$ = 0.9. While VaR remains relatively stable between the two distributions, CVaR shifts significantly with the presence of outliers, highlighting its sensitivity to tail risk.}
    \label{fig:main_cvar}
\end{figure}

For a given confidence level $\alpha$, CVaR quantifies the expected loss in the tail beyond the Value at Risk (VaR), defined as:
\begin{equation}
    \text{CVaR}_{\alpha} = \mathbb{E}[G_0|G_0\leq \text{V}_{\alpha}]
\end{equation}
Unlike VaR, which only identifies the threshold loss at a given confidence level, CVaR preserves convexity and quantifies policy performance in the bottom $1-\alpha$ percentile. This can be seen in Figure \ref{fig:main_cvar}.

As closed-form expressions for the reward distributions under the policy cannot be derived, we use Monte Carlo estimation to compute VaR and CVaR values. Since the MORSE algorithm already performs Monte Carlo rollouts to evaluate policies across $E$ episodes, we obtain the return samples at no additional computational cost. If $D = \{G_0^0, \cdots, G_0^{n_e}\}$ denotes $n_e$ independent and identically distributed return samples sorted in ascending order, then VaR can be defined as:
\begin{equation}
    V_\alpha (D)\approx G_0^{[K(1-\alpha)]}
\end{equation}
where $[n_e(1-\alpha)]$ is the index corresponding to the $\alpha-$percentile of the sorted returns. This provides an approximation of the threshold below which the returns fails with probability $1-\alpha$. Therefore, to estimate CVaR, we compute the average of the returns that fall below the VaR threshold: 
\begin{equation}
    CVaR_\alpha (D)\approx V_\alpha(D) + \frac{1}{[n_e(1-\alpha )]}\sum_{i=1}^{n_e}[G_0^i -V_\alpha (D)]^-
\end{equation}
Here, the notation $[x]^- = \text{max}(-x,0)$ denotes the negative part of a value, so the summation accumulates only the losses - i.e., return below the VaR level. This effectively captures the expected shortfall in the tail of the distribution, giving a more conservative and risk-sensitive performance metric. 

In our modified framework, CVaR is used to evaluate the reward functions of policies during the evolutionary process, rather than being treated as an additional objective. For each individual policy in the population, CVaR is calculated for each objective function using empirical distributions derived from 500 sampled episodes. This risk-aware evaluation is then incorporated into the NSGA-II algorithm to guide the optimization process and ensure hat the Pareto front reflects both performance and risk considerations. The modification can be seen in Algorithm \ref{alg:cvar}.
\begin{algorithm}
\caption{Risk-Aware MORSE}
\label{alg:cvar}
\begin{algorithmic}[1]
\State \textbf{Input:} Number of policies $n_\pi$, Maximum generations $n_g$, Evaluation episodes $n_e$, Confidence level $\alpha$
\State \textbf{Output:} Pareto front set of policies $\mathcal{F}_{\text{Pareto}}$
\State \textbf{Step 1: Initialization}
\State Generate a population of policies $\Pi = \{ \pi_1, \pi_2, \ldots, \pi_{n_\pi}\}$, where each policy $\pi_i$ is parameterized by $\mathbf{\theta}_i$ \Comment{Initialize population}

\State \textbf{Step 2: Policy Evaluation}
\For{each policy $\pi_i \in \Pi$}
    \State Initialize list of episodic rewards for each objective: $R_{\text{episodic}}^j = []$ for each objective $j$ \Comment{Prepare to store episodic rewards}
    \For{each episode $e = 1, 2, \ldots, n_e$}
        \State Initialize state $\mathbf{s}_0$ from the environment \Comment{Reset environment for new episode}
        \State Initialize episode rewards: $R^j_e = 0$ for each objective $j$
        \For{each time step $t = 0, 1, \ldots, t_{tot}$}
            \State Select action $\mathbf{a}_t \sim \pi_{\theta_i}(\cdot | \mathbf{s}_t)$ based on the policy \Comment{Policy inference}
            \State Execute action $\mathbf{a}_t$, observe reward $r_t$ and next state $\mathbf{s}_{t+1}$ \Comment{Environment interaction}
            \State Accumulate discounted reward for each objective $j$:
            \State $R^j_e \gets R^j_e + \gamma^t r_t^j$ \Comment{Calculate episode reward}
        \EndFor
        \State Append $R^j_e$ to $R_{\text{episodic}}^j$ for each objective $j$ \Comment{Store reward for the current episode}
    \EndFor
    \State \textcolor{blue}{\textbf{Step 2.1: Calculate CVaR for each objective function}}
    \For{\textcolor{blue}{each objective $j$ from $1$ to $n_f$}} \Comment{$n_f$ is the number of objective functions}
        \State \textcolor{blue}{Sort the $R_{\text{episodic}}^j$ values across the $n_e$ episodes in ascending order:} 
        \State \textcolor{blue}{$R^j_{(1)}, R^j_{(2)}, \ldots, R^j_{(n_e)}$}
        \State \textcolor{blue}{Compute the Value-at-Risk (VaR) for objective $j$ at confidence level $\alpha$:} \Comment{VaR is the $\lceil n_e(1 - \alpha) \rceil$-th smallest reward}
        \State \textcolor{blue}{$V_\alpha^j = R^j_{(\lceil n_e(1 - \alpha) \rceil)}$}
        \State \textcolor{blue}{Compute the Conditional Value-at-Risk (CVaR) for objective $j$:} \Comment{Average of the $\lceil n_e(1 - \alpha) \rceil$ worst rewards}
        \State \textcolor{blue}{$CVaR_\alpha^j = \frac{1}{\lceil n_e(1 - \alpha) \rceil} \sum_{l=1}^{\lceil n_e(1 - \alpha) \rceil} \left( R^j_{(l)} \right)$}
    \EndFor
    \State \textcolor{blue}{Store the calculated CVaR values for the policy: $\mathbf{f}_i = [CVaR_\alpha^1, CVaR_\alpha^2, \ldots, CVaR_\alpha^{n_f}]$} \Comment{Policy's fitness vector is its CVaR values}
\EndFor
\State \textbf{Step 3: Non-dominated Sorting}
\State Sort policies into fronts based on dominance relationships:
\State $\mathcal{F}_1, \mathcal{F}_2, \ldots \mathcal{F}_K$ \Comment{Apply non-dominated sorting algorithm}

\State \textbf{Step 4: Crowding Distance Calculation}
\For{each front $\mathcal{F}_k$}
    \State Compute crowding distance $d(\pi_{\theta_i})$ for each policy $\pi_{\theta_i}$:
    \State $d(\pi_{\theta_i}) = \sum_{j=1}^{n_f} \left( \frac{f_{j, \text{next}} - f_{j, \text{prev}}}{f_{j, \text{max}} - f_{j, \text{min}}} \right)$ \Comment{Calculate crowding distance for diversity}
\EndFor

\State \textbf{Step 5: Selection and Reproduction}
\State Perform binary tournament selection based on rank and crowding distance, followed by crossover and mutation:
\State $\theta_{\text{offspring}} = \text{Crossover}(\theta_{i}, \theta_{l}) + \text{Mutation}(\theta_{i}')$ \Comment{Generate new policies}

\State \textbf{Step 6: Survival Selection}
\State Combine parent population $\mathcal{P}$ and offspring $\mathcal{P}_{\text{offspring}}$, then select top $N$ policies based on non-domination rank and crowding distance:
\State $\mathcal{P}' = \text{Top-}N(\mathcal{P} \cup \mathcal{P}_{\text{offspring}})$ \Comment{Select next generation population}

\State \textbf{Step 7: Termination Criteria}
\If{Termination criteria met (e.g., $g \geq G$)}
    \State \textbf{break} \Comment{Exit loop if criteria satisfied}
\EndIf

\State \textbf{Step 8: Pareto Front Identification}
\State Identify non-dominated solutions in the final population:
\State $\mathcal{F}_{\text{Pareto}} = \{ \pi_i : \pi_i \text{ is non-dominated in } \mathcal{P}' \}$ \Comment{Extract final Pareto front}

\State \textbf{Return} Pareto front set $\mathcal{F}_{\text{Pareto}}, \quad \Pi^* = \{\pi_1^*, \dots, \pi_{n_{\pi}^*}^*\}$ \Comment{Return optimal policies}
\end{algorithmic}
\end{algorithm}

\begin{figure}[htbp]
    \centering
        \begin{subfigure}[b]{0.32\textwidth}
        \centering
        \includegraphics[width=\textwidth]{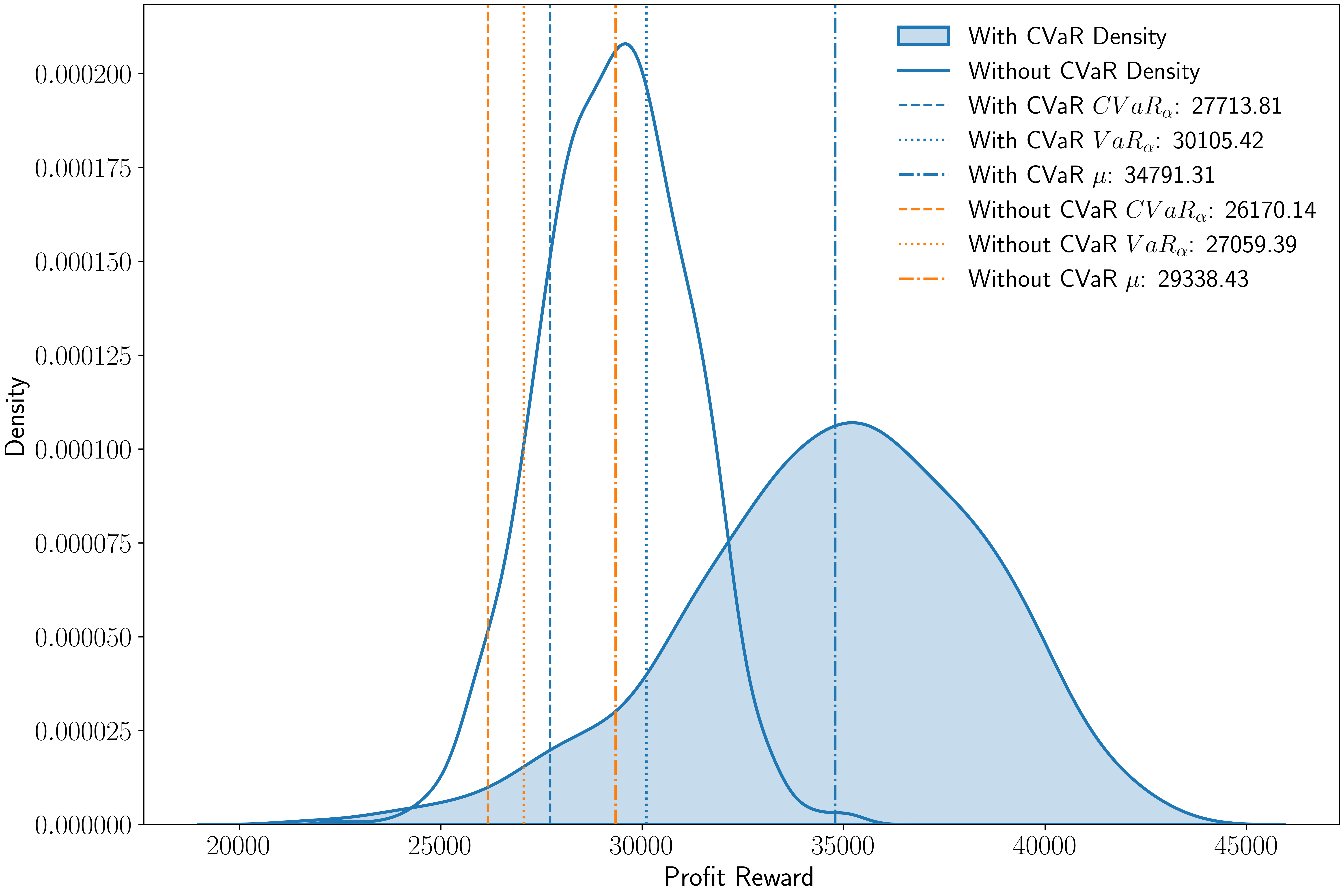}
        \caption{Cumulative Profit Reward Distribution}
        \label{fig:pro_dist}
    \end{subfigure}
    \hfill
    \begin{subfigure}[b]{0.32\textwidth}
        \centering
        \includegraphics[width=\textwidth]{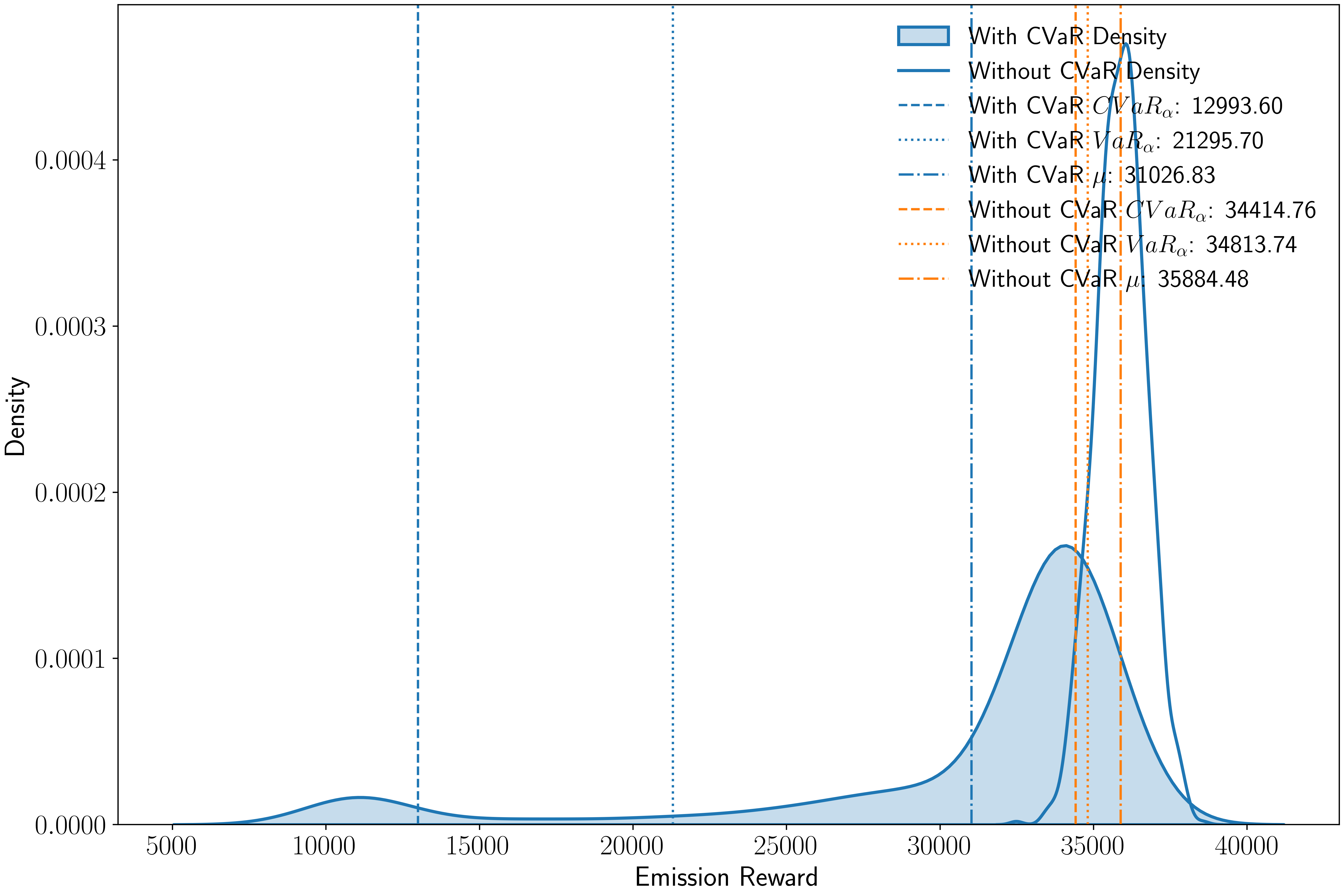}
        \caption{Emissions Reward Distribution}
        \label{fig:emi_dist}
    \end{subfigure}
    \hfill
    \begin{subfigure}[b]{0.32\textwidth}
        \centering
        \includegraphics[width=\textwidth]{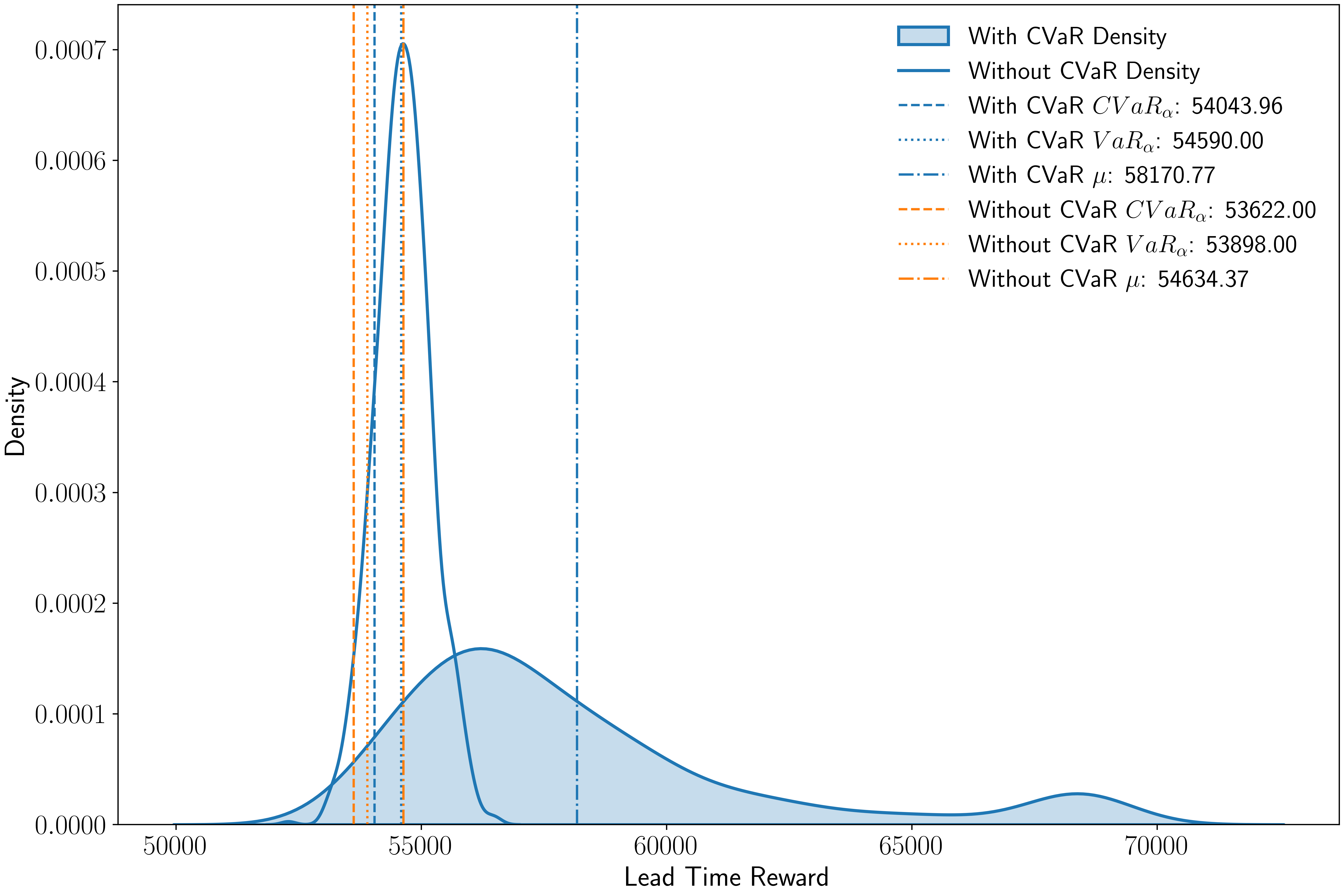}
        \caption{Lead Time Reward Distribution}
        \label{fig:lt_dist}
    \end{subfigure}
    
    \caption{Overlayed reward distributions for CVaR-trained and mean-trained policies across 1,000 Monte Carlo simulations. Density plots show the spread and skewness of rewards and are annotated with CVaR, VaR, and mean values to facilitate direct comparison of risk-averse versus risk-neutral behavior.}
    \label{fig:cvar_results}
\end{figure}

Once the Pareto front has been trained, we randomly select a policy from the set of CVaR-trained solutions and compare it to a policy from the set of non-CVaR-trained solutions. To evaluate and visualize the performance of these policies, we run 1,000 Monte Carlo simulations for each selected policy. The goal is to observe and compare the reward distributions across the CVaR-trained policies compared to the non-CVaR-trained ones as seen in Figure \ref{fig:cvar_results}. 

From Figure \ref{fig:cvar_results}, the overlayed reward distributions for CVaR-trained and mean-trained policies provide a direct comparison of risk-averse and risk-neutral behaviors across 1,000 Monte Carlo simulations. Conditional Value at Risk (CVaR) measures the average of the worst-case outcomes within a distribution, providing a quantitative assessment of tail risk. For minimization objectives such as emissions or lead time, the worst outcomes correspond to the highest values. Therefore, by explicitly optimizing for CVaR, the CVaR-trained policies reduce these extreme high values, resulting in a lower CVaR relative to mean-trained policies, even though the the mean of the distribution may increase, decrease, or remain approximately the same, depending on the trade-offs made during optimization. For maximization objectives such as profit, the worst outcomes are the lowest values, therefore CVaR-trained policies improve the worst-case outcomes, leading o a higher CVaR than than mean-trained policies. In both minimization and maximization contexts, the key indicator of success is the improvement of the CVaR relative to the mean-trained baseline, rather than changes in the mean itself.

These differences are particularly important in stochastic environments where rare but extreme events can dominate system performance. By controlling tail risk, CVaR-trained policies provide more predictable and stable outcomes, reducing the likelihood of catastrophic scenarios. The overlayed distributions confirm that, for all objectives, CVaR-trained policies achieve the intended tail improvements: for minimization objectives, CVaR decreases relative to the mean-trained policy, and for maximization objectives, CVaR increases. This demonstrates that the methodology successfully balances risk and performance, capturing both risk-neutral and risk-averse behaviors within a single framework, and highlights its applicability across diverse decision-making environments including supply chains and process systems engineering.

\section{Benchmarking}\label{benchmark}
To evaluate the performance and effectiveness of our proposed method, we conduct a benchmarking analysis against two well-established algorithms: Concave-Augmented Pareto Q-Learning (CAPQL) \citep{lu2023multi} and Multi-Objective Neural Evolution Strategy (MONES) \citep{shao2021multi}. These algorithms represent state-of-the-art approaches in the field of multi-objective reinforcement learning (MORL) and are selected for comparison based on their ability to handle continuous action spaces and multi-objective optimization problems. We assess their performance across all three objectives considered in our problem formulation to provide a thorough evaluation of their strengths and limitations in relation to our approach.

\begin{figure}[h!]
    \centering
    \includegraphics[width=\linewidth]{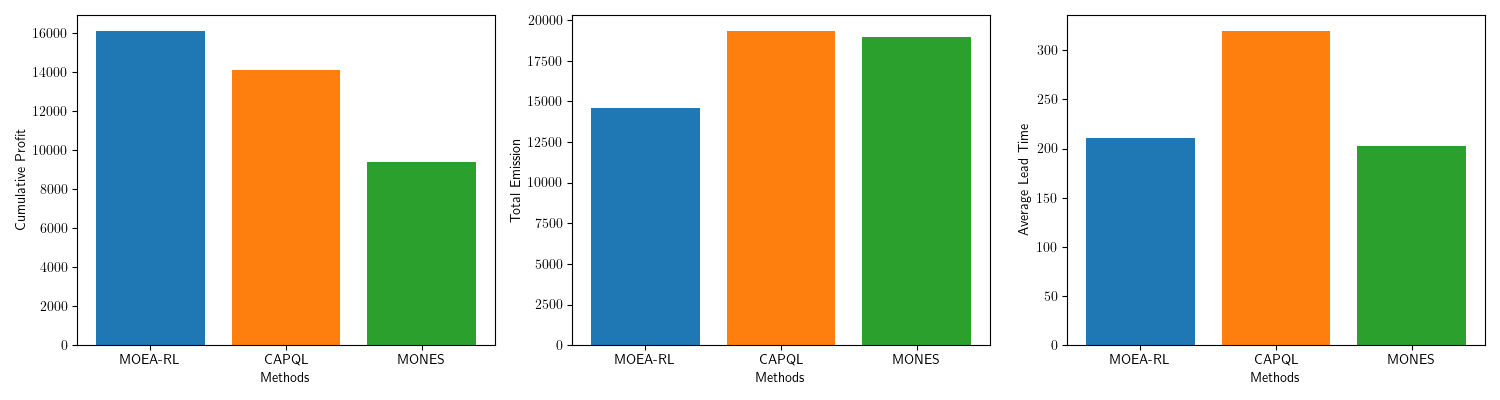}
    \caption{ Performance comparison of our methodology against other MORL methods: CAPQL and MONES across the three different reward functions (profit, emissions, lead time)}
    \label{fig:benchmark}
\end{figure}

As seen in Figure \ref{fig:benchmark}, our methodology outperforms both CAPQL and MONES, achieving superior performance across objectives in the inventory management case study presented in Section \ref{sec:case_study}. This analysis highlights the advantages of our approach, particularly in optimizing complex, multi-objective systems.

The superior performance of our approach, as compared to CAPQL and MONES, can be attributed to several factors, with the use of evolutionary algorithms (EAs) being a significant contributor. EAs are particularly effective in solving multi-objective optimization problems because they inherently support the exploration of diverse solutions, which helps prevent the optimization process from getting trapped in local optima. In contrast, approaches like CAPQL and MONES rely more heavily on gradient-based optimization, which can suffer from slow convergence or local optima, especially in non-convex spaces such as those encountered in complex inventory management problems.

Furthermore, previous research has shown that Evolutionary Algorithm-based Reinforcement Learning (EA-RL) approaches, including those by OpenAI \citep{salimans2017evolution}, have demonstrated superior or comparable performance in various complex tasks compared to traditional gradient-based methods. This includes applications across multi-objective optimization problems, where EA-RL excels at balancing exploration and exploitation to discover more diverse and robust solutions.

\section{Conclusion and Future Work} \label{conc}
In this paper, we have proposed a strategy that integrates Reinforcement Learning (RL) with Multi-Objective Evolutionary Algorithms (MOEAs) to address the complex and dynamic nature of supply chain management. We demonstrate that using MOEAs to explore the policy neural network parameter space results in a Pareto front of policies, providing the decision-maker with a set of policies that can be dynamically switched based on the current system objectives. This capability enhances fast decision-making, resilience, and flexibility in uncertain and evolving environments. Our approach is particularly effective in handling disruptions, showcasing its adaptability and ability to balance multiple conflicting objectives in real-time. We introduce Conditional Value-at-Risk (CVaR) as a mechanism to incorporate risk-sensitive decision-making, showcasing the flexibility of our framework for more risk-averse decision strategies. This addition enhances the robustness of the policies and ensures that they not only optimize performance but also account for the risks associated with extreme outcomes. Finally, we validate our method through a series of case studies and benchmarking experiments, demonstrating its potential to optimize complex, multi-objective problems commonly encountered in supply chain management.

For future work, we plan to enhance the proposed approach by integrating human expertise to improve the search efficiency of the evolutionary algorithms, enabling more informed exploration of the solution space. Additionally,  we also plan to expand our method to multi-agent settings, allowing for collaborative decision-making within supply chains, which could lead to more effective coordination and optimization across multiple agents.

The codes are available at: \href{https://github.com/nikikotecha/MORSE}{github.com/nikikotecha/MORSE}

\section*{Acknowledgements}
Niki Kotecha acknowledges support from I-X at Imperial College London. 

\bibliographystyle{elsarticle-harv} 
\bibliography{ref}

\newpage
\end{document}